\documentclass[10pt,twocolumn,letterpaper]{article}

\usepackage{iccv}              % To produce the CAMERA-READY version
\usepackage[accsupp]{axessibility}
\usepackage{amsmath}
\usepackage{booktabs}
\usepackage{multirow}
\usepackage{makecell}
\usepackage{stfloats}
\usepackage{amsthm}

\newcommand{\insideskip}{\vspace{-1ex}}
\newcommand{\outsideskip}{\vspace{-1ex}}

\newtheorem{lemma}{Lemma}

\definecolor{iccvblue}{rgb}{0.21,0.49,0.74}
\usepackage[pagebackref,breaklinks,colorlinks,allcolors=iccvblue]{hyperref}

\def\paperID{11217} % *** Enter the Paper ID here
\def\confName{ICCV}
\def\confYear{2025}

\title{Diffusion Epistemic Uncertainty with Asymmetric Learning for Diffusion-Generated Image Detection}

\author{
Yingsong Huang$^{1}$\thanks{These authors contributed equally to this work.}\;, Hui Guo$^{1}$\footnotemark[1]\;, Jing Huang$^{2}$\footnotemark[1]\;, Bing Bai$^{3}$\thanks{Corresponding authors.}\; and Qi Xiong$^{1}$\footnotemark[2]\\
$^{1}$Tencent Inc.\quad $^{2}$Hikvision\quad $^{3}$Microsoft MAI\\
{\tt\small \{hudsonhuang,emmaguo\}@tencent.com, huangjing29@hikvision.com}\\
{\tt\small bingbai@microsoft.com, keonxiong@tencent.com}
}

\begin{document}
\maketitle
\begin{abstract}
% The accelerated advancement of diffusion models highlights the imperative for the detection of generated images. 
% Existing research has shown that using diffusion measurements like reconstruction error enhances the generalizability of detectors.
% However, it is crucial to acknowledge the presence of aleatoric and epistemic uncertainty in the diffusion process, as they have distinct impacts on the effectiveness of the reconstruction error. 
% Aleatoric uncertainty, which captures inherent data noise, introduces predictive ambiguity that impedes the identification of generated images.
% In contrast, epistemic uncertainty, which accounts for the model’s ignorance of unfamiliar patterns, helps to identify generated images.
% In this paper, we propose a novel framework, Diffusion Epistemic Uncertainty with Asymmetric learning~(DEUA) for diffusion-generated images detection. 
% We introduce a new representation, referred to as Diffusion Epistemic Uncertainty~(DEU), to determine whether data is close to or far away from the manifold of diffusion-generated samples.
% We utilize the Laplace approximation to quantify epistemic uncertainty in the diffusion process. 
% Additionally, we adopt an asymmetric loss function to train a balanced classifier with larger margins, further enhancing generalizability.
% The results on large-scale benchmarks demonstrate the effectiveness of the proposed method. 
The rapid progress of diffusion models highlights the growing need for detecting generated images.
Previous research demonstrates that incorporating diffusion-based measurements, such as reconstruction error, can enhance the generalizability of detectors. 
However, ignoring the differing impacts of aleatoric and epistemic uncertainty on reconstruction error can undermine detection performance.
Aleatoric uncertainty, arising from inherent data noise, creates ambiguity that impedes accurate detection of generated images. 
As it reflects random variations within the data (e.g., noise in natural textures), it does not help distinguish generated images. 
In contrast, epistemic uncertainty, which represents the model's lack of knowledge about unfamiliar patterns, supports detection.
In this paper, we propose a novel framework, Diffusion Epistemic Uncertainty with Asymmetric Learning~(DEUA), for detecting diffusion-generated images.
We introduce Diffusion Epistemic Uncertainty~(DEU) estimation via the Laplace approximation to assess the proximity of data to the manifold of diffusion-generated samples.
Additionally, an asymmetric loss function is introduced to train a balanced classifier with larger margins, further enhancing generalizability.
Extensive experiments on large-scale benchmarks validate the state-of-the-art performance of our method.
\end{abstract}    
\section{Introduction}
\label{sec:intro}

The rapid advancement of diffusion models~\cite{ho2020denoising,song2020denoising,Rombach_Blattmann_Lorenz_Esser_Ommer_2022,Dhariwal_Nichol_2021,gu2022vector,Nichol2021GLIDETP} has enabled the generation of images that are nearly indistinguishable from real ones to the human eye. 
This raises concerns about potential misuse of generated images in various fields, such as security risks and privacy violations~\cite{juefei2022countering}. 
Consequently, there is an increasing need for effective methods to detect diffusion-generated images.

% Numerous studies~\cite{tan2024rethinking,corvi2023intriguing,corvi2023detection,wang2023dire,ma2023exposing,luo2024lare,zhang2023diffusion} have been conducted to detect and analyze the characteristics of images generated by diffusion models.
% Accordingly, visual artifacts~\cite{tan2024rethinking,ojha2023towards} and spectrum artifacts~\cite{corvi2023intriguing,corvi2023detection} have been used to distinguish generated images.
% However, these representations are constructed on the basis of observations of the image content and cannot reveal the differences between real and fake images in terms of data distribution.
% Therefore, these representations suffer limited generalization to unseen diffusion-generated images.
%？lare2也使用了clip与dire结合，method里面直接follow这种方式
%？我们claim LUD的表征比DIRE好，但是只能从pixel distribution上说明，但是单独的分类效果却不好
%但是，这些representation是从图像内容出发被构造出来，并不能在数据分布上揭示真伪图像的差异。
%因此，很难被深度网络学到一个general分类器。

%此处是否要补充
%以上的方法存在什么问题。
%我们的方法分为2个部分：1、从数据分布的角度xxx。2、学习一个均匀的分类界面

% Recently, diffusion measurements, such as the reconstruction error~\cite{wang2023dire,ma2023exposing,luo2024lare}, have been proposed as a discriminative feature for diffusion-generated image detection.
Recent approaches have proposed using diffusion measurements, such as reconstruction error~\cite{wang2023dire,ma2023exposing,luo2024lare}, as discriminative features for detecting diffusion-generated images.
% and shows great generalizability across different diffusion models.
These methods assume that images generated by the diffusion model and their corresponding reconstructions share the same distribution. As a result, pre-trained diffusion models can reconstruct generated images more accurately, whereas the reconstruction error for real images is typically larger.
% Based on this assumption, the image reconstruction error and latent reconstruction error are used as features to distinguish forged images from real images.
%最近的一系列工作开始从数据分布的角度提出
%Aleatoric uncertainty and epistemic uncertainty can then be used to induce predictive uncertainty, the confidence we have in a prediction. cannot be explained away
%然而，在扩散过程中存在两种uncertainty。其中aleatoric uncertainty会造成现有diffusion特征的ambiguity（注意是会减弱这些特征对ood samples的区分能力，而不是完全失去检测能力），而epistemic则能区分ood samples。
%我们对diffusion中的uncertainty进行解耦，单独提取其中的epistemic部分来进行ood detection以达到deepfake detection。
%后面的图示，均是对于dire的图例，即image space的，对于latent space（LaRE）的图例有类似的结论，但是放在appendix。（因为我们发现vaedecoder能进一步增加差异？）

% However, there exists uncertainty~(i.e., aleatoric uncertainty and epistemic uncertainty) in the diffusion process, and the diffusion model may produce misleading predictions due to inherent uncertainty~\cite{chan2024hyper}.
%然而，diffusion reconstruction error, 并不仅仅因为ood才会增加
%diffusion process中存在两种uncertianty，他们以不同的方式影响/增加 reconstruction error.
However, large reconstruction errors are not solely caused by out-of-distribution~(OOD) data. Both aleatoric and epistemic uncertainty, inherent in the diffusion~\cite{chan2024hyper}, contribute to reconstruction error. Aleatoric uncertainty reflects the inherent noise and randomness in the data. As it remains relatively constant across both in-distribution and OOD samples~\cite{kendall2017uncertainties}, it does not significantly aid in detecting diffusion-generated images. This insight motivates a shift from reconstruction error towards epistemic uncertainty.

\begin{figure}[t]
  \centering
  \begin{subfigure}{0.32\linewidth}
    \includegraphics[width=1.0\columnwidth]{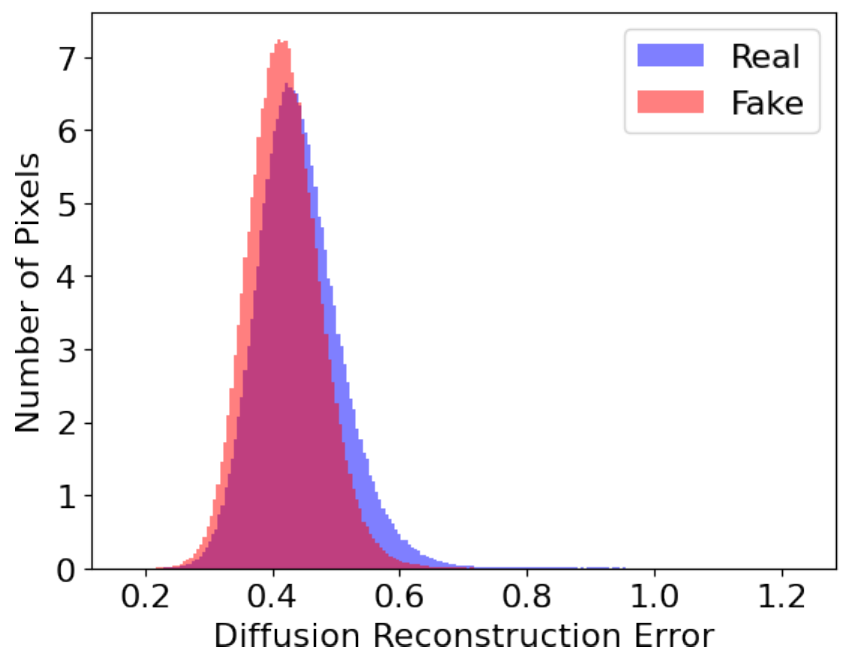}   
    \caption{Distribution of reconstruction error.}
    \label{fig:re}
  \end{subfigure}
  \hfill
  \begin{subfigure}{0.32\linewidth}
 \includegraphics[width=1.0\columnwidth]{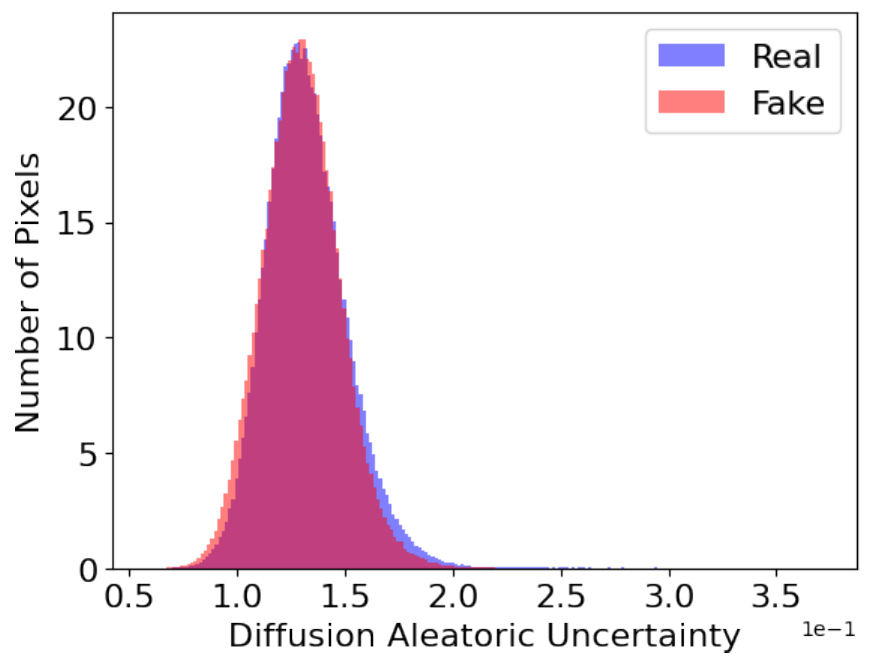}
    \caption{Distribution of aleatoric uncertainty.}
    \label{fig:au}
  \end{subfigure}
  \hfill
  \begin{subfigure}{0.32\linewidth}
 \includegraphics[width=1.0\columnwidth]{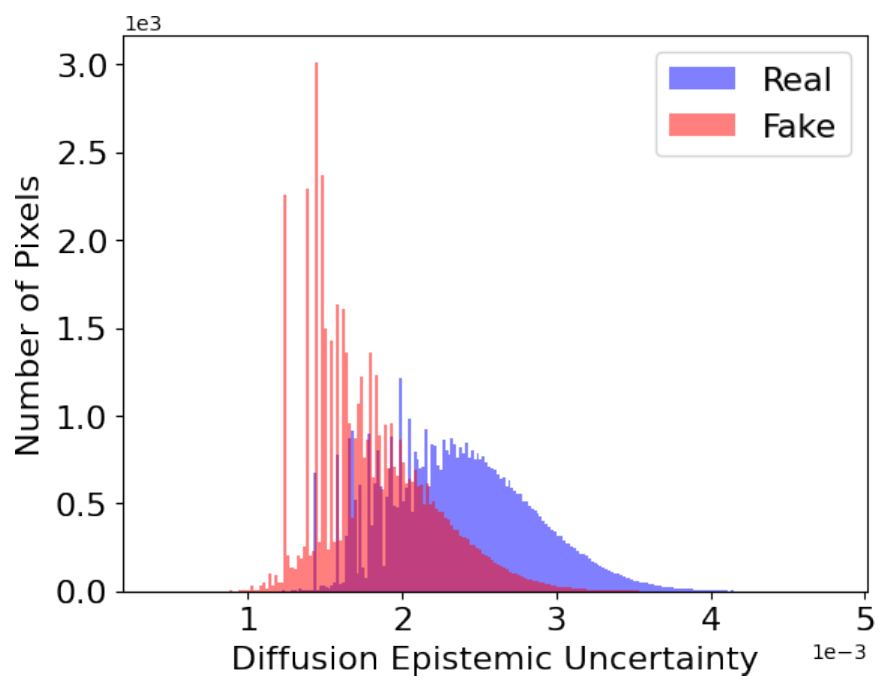}
    \caption{Distribution of epistemic uncertainty.}
    \label{fig:eu}
  \end{subfigure}
\insideskip
  \caption{Distribution of diffusion reconstruction error, aleatoric uncertainty, and epistemic uncertainty in real and generated samples. (a)~The reconstruction error distribution for real samples overlaps with that of fake samples due to the presence of aleatoric uncertainty. (b)~The distribution of aleatoric uncertainty is nearly indistinguishable between real and fake samples. (c)~Epistemic uncertainty more accurately distinguishes real samples from fake ones.}
  \label{fig:dist}
\outsideskip
\end{figure}

Epistemic uncertainty, in contrast, captures the model's lack of knowledge about unfamiliar patterns and increases significantly for samples that fall outside the domain of the training data~\cite{kendall2017uncertainties,maddox2019simple,deng2021libre,DBLP:conf/aaai/HuangBZBW22}. Consequently, epistemic uncertainty provides a more reliable basis for anomaly detection. \Cref{fig:dist} demonstrates that epistemic uncertainty more accurately distinguishes real samples from fake ones.
Therefore, for detecting generated images, it is essential to disentangle the overall uncertainty in diffusion measurements and focus specifically on epistemic uncertainty.

Based on this insight, we propose Diffusion Epistemic Uncertainty with Asymmetric Learning~(DEUA), a novel framework for detecting diffusion-generated images. Our framework separates overall uncertainty into aleatoric and epistemic components, using Diffusion Epistemic Uncertainty~(DEU) to identify samples close to the manifold of diffusion-generated images. We employ the last-layer Laplace approximation~\cite{kristiadi2020being,daxberger2021laplace} to estimate epistemic uncertainty in the diffusion process.
%we estimate unceraint in the latent space.
% We find that uncertainty estimation with a single-step sample is enough to distinguish the real images from the diffusion-generated images.
%然而，定式深度网络预测结果本身不能表示预测的不确定性\cite{},因此reconstruction error may be not trustworthy for ood inputs
%如图pixel distri
%于是，我们提出LUD直接反映数据分布的差异
%如图
%我们提出xxx来估计LCD
%(a)是否替换为latent dire？
% \begin{figure}[ht]
% \vskip 0.2in
% \begin{center}
% \subfigure[Distribution of reconstruction error]{
%         \label{Fig1.a}
%         \includegraphics[width=0.46\columnwidth]{pics/re.pdf}    
%     }
%     \subfigure[Distribution of epistemic uncertainty]{
%         \label{Fig1.b}
%         \includegraphics[width=0.46\columnwidth]{pics/uq.pdf}
%     }
% \caption{Diffusion reconstruction error and diffusion epistemic uncertainty distribution on real and diffusion-generated images.(a)~Reconstruction error distribution of the real samples overlaps with the fake samples due to ambiguous predictions. (b)~Epistemic uncertainty distinguishes the real samples from the fake ones more accurately.}
% \label{Fig1}
% \end{center}
% \vskip -0.2in
% \end{figure}
%当我们使用DEU训练分类模型是，我们也遇到了sink class现象
%为了充分训练泛化特征，我们使用了非对称对比学习损失。
% When we exploit DEU to train a classification model, we meet a phenomenon similar to the `sink class' with a asymmetric separation of real and fake classes~\cite{ojha2023towards}.
% To solve this problem, we introduce an asymmetric loss to maximize the preservation of information content of the representation and rectify the decision boundary.
Moreover, during the training of a classifier, we observe a phenomenon akin to the  ``sink class'' concept, where there is asymmetric separation between the real and fake classes~\cite{ojha2023towards}. 
To address this, we introduce an asymmetric contrastive loss function that aims to train a classifier with a balanced decision boundary and larger margins, thus improving its generalizability.

The key contributions of this work are summarized as follows:
\begin{itemize}
% \item We analyze why utilizing the diffusion reconstruction error in existing methods is insufficient for generated images detection and scribe
% the reasons for predictive ambiguity caused by aleatoric uncertainty.
% \item We propose a novel framework for diffusion-generated images detection. 
% The framework exploits the diffusion epistemic uncertainty to capture the underlying patterns in the data and leverage asymmetric contrastive learning to learn a classification model of generalizability.
\item We analyze why the use of diffusion reconstruction error in existing methods is insufficient for detecting generated images and identify the predictive ambiguity introduced by aleatoric uncertainty.
\item We introduce a novel framework for detecting diffusion-generated images, leveraging diffusion epistemic uncertainty to identify samples close to the manifold of generated images, and using asymmetric contrastive learning to train a classifier with enhanced generalizability.
\item Extensive experiments on large-scale datasets, namely GenImage and DRCT-2M, validate the effectiveness of the proposed framework.
\end{itemize}
%trained on sd1.4 and test on other diffusion models
%reconstruction因为gaussain noise引入了Aleatoric uncertainty
%受Aleatoric uncertainty影响，recontruction error不能反应ood的差异
%Aleatoric uncertainty cannot be explained away with more data,
%Aleatoric uncertainty does not increase for out-of-data examples (situations different from training set), whereas epistemic uncertainty does.
%~\cite{kendall2017uncertainties}
%因为Aleatoric所以受image content frequency的影响
%怎么解释Epistemic Uncertainty仍需要clip feature
%这里的核心似乎在说Epistemic Uncertainty不受image content frequency的影响？或者影响小？
\section{Related Work}
\label{sec:related}
In this section, we review related work on generalizable generated image detection and uncertainty estimation in diffusion.

\subsection{Generalizable Generated Image Detection}
In recent years, research on detecting AI‑generated images has evolved from hunting for obvious artifacts to probing increasingly subtle statistical cues. Early studies targeted visible flaws—including abnormal color \cite{mccloskey2018detecting}, lighting inconsistencies \cite{farid2022lighting}, texture anomalies \cite{liu2020global}, and blending seams \cite{li2020face}—but such errors have largely disappeared as generators improve. Accordingly, the community shifted to fine‑grained spatial \cite{gragnaniello2021gan,tan2023learning,chandrasegaran2022discovering} and spectral signatures \cite{zhang2019detecting,frank2020leveraging,qian2020thinking,tan2024frequency}, work that was initially developed for GAN imagery \cite{lin2024detecting}. Subsequent studies revealed that diffusion models also leave tell‑tale spectral traces \cite{corvi2023intriguing,corvi2023detection}, while NPR \cite{tan2024rethinking} exploited local pixel dependencies created by up‑sampling modules, capturing both GAN and diffusion fakes. To bolster cross‑model generalization, several approaches embed or fuse local forgery evidence within CLIP’s vision‑language space \cite{ojha2023towards,sha2023fake,cozzolino2024raising} or jointly reason over spatial–frequency domains, as in FatFormer and AIDE \cite{liu2024forgery,yan2024sanity}. 
A parallel line of work leverages the pretrained diffusion model itself: DIRE \cite{wang2023dire} and SeDID \cite{ma2023exposing} rely on reconstruction error, DNF \cite{zhang2023diffusion} analyzes predicted noise during inversion, and LaRE$^{2}$ \cite{luo2024lare} achieves similar benefits with a single‑step latent error, all building on the premise that synthetic images are more faithfully reconstructed than real ones. We argue that randomness in the forward diffusion process introduces prediction ambiguity and instead measure \emph{epistemic uncertainty}—how unsure the model is about its own predictions—to gauge a sample’s proximity to the synthetic manifold. Concurrently, multimodal large models such as ForgeryGPT \cite{liu2024forgerygpt} and HEIE \cite{yang2025heie} provide interpretable detection via cross‑modal or chain‑of‑thought reasoning, and a new wave of zero‑/few‑shot methods (e.g., ZED \cite{cozzolino2024zero}, FSD \cite{wu2025few}) achieves robustness to unseen generators without extensive synthetic data. 

Against this backdrop, our work delves into the generation process itself and introduces a diffusion‑based epistemic‑uncertainty metric that complements reconstruction‑ and entropy‑based criteria, offering a principled, source‑agnostic signal for detecting AI‑generated images.

\subsection{Uncertainty Estimation and Laplace Approximation}
%Uncertainty estimation to build confidence in predictions is crucial for deep learning and has been intensively studied~\cite{lakshminarayanan2016simple,gal2016uncertainty,maddox2019simple}.
Probabilistic modeling methods, such as Bayesian Neural Networks~(BNNs)~\cite{gal2016uncertainty,maddox2019simple} and deep ensembles~\cite{lakshminarayanan2016simple} have demonstrated the ability to estimate uncertainty.
BNNs place a prior distribution over the model parameters and quantify uncertainty as the sample variance over the weight distribution. 
Deep ensembles train an ensemble of networks with different weight initializations and take the ensemble variance as a measure of uncertainty.

Due to BNNs' high nonlinearity, it is infeasible to analytically compute the Bayesian posterior of parameters.
Approximate inference techniques, such as variational inference~(VI)~\cite{blundell2015weight} and Laplace approximation~(LA)~\cite{ritter2018scalable} are introduced to establish an approximation of the posterior.
LA can be effortlessly applied to pre-trained models in a post-processing manner and has strong uncertainty quantization performance~\cite{foong2019between,daxberger2021laplace}.
%Sampling weights and forwarding different networks require high computational cost.
%Laplace approximation~(LA) has recently gained particular attention because it can apply to pre-trained models effortlessly in a post-processing manner and enjoy strong uncertainty quantification (UQ) performance

Recently, there has been some exploration around uncertainty estimation in diffusion~\cite{kou2023bayesdiff,chan2024hyper}.
Our estimation method is related to BayesDiff~\cite{kou2023bayesdiff}.
The main difference is that our method qualifies the uncertainty in the forward and reverse diffusion process on existing images while BayesDiff focuses on characterizing the uncertainty in the reverse diffusion process on images to be generated.
%我的估计方式somehow与Bayesdiff有紧密联系
%Bayesdiff只评估了reverse diffusion process的uncertainty
\section{Preliminaries}
\label{sec:preliminaries}
% Our proposed method is based on the observation of the impacts of uncertainty in the diffusion process.
In this section, we review the forward and reverse diffusion processes.
We then examine the influence of two types of uncertainty on the detection of generated images, and present case studies to illustrate these concepts.
Additionally, we briefly introduce the Laplace approximation for efficient Bayesian inference in diffusion.
We use it to estimate epistemic uncertainty in \cref{sec:method-deu}.
%因为我们的方法基于pretrained-model在diffusion process中产生的特征，所以我们先回顾diffusion forward/reverse process，讨论重建过程中的ambiguity，并给出case study。
%此外，我们简单地说明利用laplace approximation对diffusion进行bayesian inference的过程，作为后面估计方法中的基础。
\subsection{Diffusion Model}
\label{sec:pre-diffusion}
%用什么图来说明？
%1.bias/std的snr趋势图
%2.bias、std的图示
%3.即使fake dire的pixel/image均值比real dire的小，也不能说其一定有效。毕竟不能保证所有image都小
%4.是否还需要比较epsitemic uncertainty的均值标准差？不需要，figure1已经展示了其不重合
A diffusion model typically involves two processes.
The forward process defines a Markov chain of steps that gradually add Gaussian noise to the raw image $\mathbf{x}_0$ until degenerating it into isotropic Gaussian distribution, which is defined as:
\begin{equation}
    q(\mathbf{x}_t|\mathbf{x}_{t-1}) = \mathcal{N}(\mathbf{x}_t;\sqrt{\frac{\alpha_t}{\alpha_{t-1}}}\mathbf{x}_{t-1},(1-\frac{\alpha_t}{\alpha_{t-1}}\mathbf{I}))\,,
\end{equation}
where $\mathbf{x}_t$ is the noisy image at the $t$-th step, $\alpha_t$ is a predefined noise schedule, and $T$ is the total steps.
According to the property of Markov chain, we can get $\mathbf{x}_t$ from $\mathbf{x}_0$ via:
\begin{equation}
    q(\mathbf{x}_t|\mathbf{x}_0)=\mathcal{N}(\mathbf{x}_t;\sqrt{\alpha_t}\mathbf{x}_0,(1-\alpha_t)\mathbf{I})\,.
\end{equation}
The reverse process is also formulated as a Markov chain in DDPM~\cite{ho2020denoising}, using a network $p_\theta(\mathbf{x}_{t-1}|\mathbf{x}_t)$ to fit the real distribution $q(\mathbf{x}_{t-1}|\mathbf{x}_t)$:
\begin{equation}
    p_\theta(\mathbf{x}_{t-1}|\mathbf{x}_t)=\mathcal{N}(\mathbf{x}_{t-1};\boldsymbol{\mu}_\theta(\mathbf{x}_t,t),\boldsymbol{\Sigma}_\theta(\mathbf{x}_t,t))\,.
\end{equation}
In practice, the optimization target is simplified to predicting the added noise $\boldsymbol{\epsilon}$
as  another parameterization of $p_\theta(\mathbf{x}_{t-1}|\mathbf{x}_t)$.

To accelerate the iterative process, DDIM~\cite{song2020denoising} proposes a new deterministic method without the Markov hypothesis. 
The new reverse process is formulated as:
\begin{equation}
\label{xt-1}
\begin{aligned}
    \mathbf{x}_{t-1}=\sqrt{\alpha_{t-1}}(\frac{\mathbf{x}_t-\sqrt{1-\alpha_t}\boldsymbol{\epsilon}_\theta(\mathbf{x}_t,t)}{\sqrt{\alpha_t}})+\\
    \sqrt{1-\alpha_{t-1}-\sigma_t^2}\boldsymbol{\epsilon}_\theta(\mathbf{x}_t,t)+\sigma_t\boldsymbol{\epsilon}_t\,.
\end{aligned}
\end{equation}
With the formulation, the reverse process become deterministic if $\sigma_t=0$. Moreover, DDIM proposes to sample a subset of $S$ steps $\tau_1,...,\tau_S$, and thus the reverse process can generate samples following the new faster schedule.
% \subsection{Uncertainty Disambiguation in Diffusion Process}
\subsection{Rethink Uncertainty in the Diffusion Process}
\label{sec:pre-uncertainty}
% Prediction ambiguity exists, accompanied by a high level of uncertainty in the diffusion process~\cite{chan2024hyper}.
% Identifying diffusion-generated images requires disambiguation of epistemic uncertainty from aleatoric uncertainty, as they impact identification in fundamentally different ways.
Uncertainty can have a significant impact on the reconstruction error commonly used in diffusion-generated image detection~\cite{chan2024hyper}, where aleatoric and epistemic components can lead to prediction ambiguity. Distinguishing between these two forms of uncertainty is essential for accurately detecting diffusion-generated images, as they affect the detection process in fundamentally different ways.

\begin{figure}[t]
\centering
\centerline{\includegraphics[width=0.75\columnwidth]{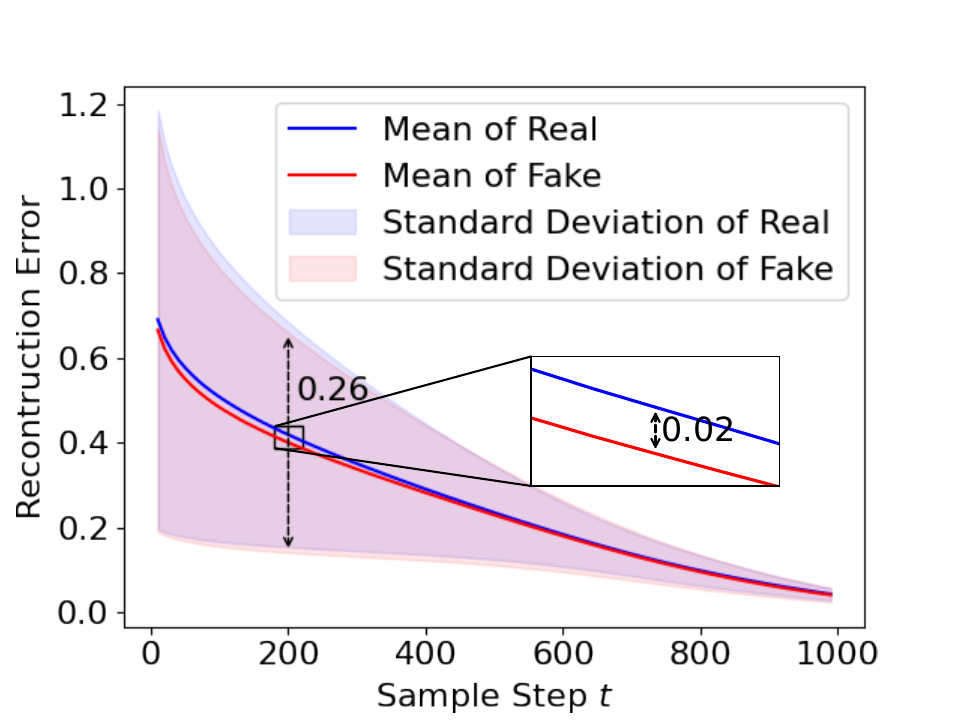}}
\insideskip
\caption{Diffusion Reconstruction error and predictive ambiguity~(1000 images are used) due to aleatoric uncertainty. The difference in reconstruction error
becomes less significant when compared to the standard
deviation caused by aleatoric uncertainty.}
\outsideskip
\label{re_uncertainty}
\end{figure}

% On the one side, aleatoric uncertainty captures the inherent noise in observed data and reflects aspects of the task which are inherently difficult~\cite{kendall2017uncertainties,gal2016uncertainty}. 
% For instance, local image patches with high-frequency details and high-level natural texture noise may exhibit increased aleatoric uncertainty during the diffusion reconstruction process.
% Notably, aleatoric uncertainty does not increase for OOD samples, and thus cannot be used to detect such samples.
% Furthermore, when using a measurement in the diffusion process~(e.g., the reconstruction error) to detect diffusion-generated images, the prediction ambiguity due to aleatoric uncertainty can lead to misleading results.
% On the other side, epistemic uncertainty accounts for the ignorance about which model generated the collected data.
% Epistemic uncertainty increases considerably for samples that lie far from the domain of training sets~\cite{kendall2017uncertainties}, making it a valuable indicator for identifying samples generated by diffusion models.
Aleatoric uncertainty arises from inherent noise in the observed data and reflects aspects of the task that are intrinsically difficult~\cite{kendall2017uncertainties,gal2016uncertainty}. For instance, image patches with high-frequency details or complex natural textures often exhibit greater aleatoric uncertainty during the diffusion reconstruction process. Notably, aleatoric uncertainty does not increase for out-of-distribution~(OOD) samples, making it unsuitable for detecting such anomalies. Moreover, relying on diffusion-based measurements (e.g., reconstruction error) to identify generated images can become misleading when aleatoric uncertainty dominates, as it inflates prediction ambiguity.
\Cref{re_uncertainty} illustrates how aleatoric uncertainty causes predictive ambiguity in reconstruction error estimates. Specifically, the reconstruction error is computed as the mean of sampled diffusion errors, while aleatoric uncertainty is estimated via the sample variance. To ensure robust statistics, we use a set of 1,000 images, each sampled 20 times, and take the mean across all pixels. Although there is a notable gap between the mean reconstruction errors of real and fake images, the standard deviation introduced by aleatoric uncertainty diminishes this difference. Consequently, as shown in \cref{fig:re}, relying solely on reconstruction error is less effective for distinguishing real from fake images when aleatoric uncertainty is high.

% \Cref{re_uncertainty} presents the issue of predictive ambiguity due to aleatoric uncertainty in the diffusion reconstruction error. 
% The reconstruction error is computed as the mean over the set of predicted errors sampled from diffusion process, while aleatoric uncertainty is captured by the sample variances.
% To ensure stable statistical outcomes, we use a set of 1000 images and sample each 20 times. 
% Both the reconstruction error and aleatoric uncertainty are calculated as the mean across all pixels in these 1000 images.
% As depicted in the figure, there is a discernible gap between the mean reconstruction errors of real and fake images. 
% However, this difference becomes less significant when compared to the standard deviation due to aleatoric uncertainty. 
% The presence of aleatoric uncertainty significantly reduces the effectiveness of the reconstruction error in distinguishing between real and fake images, as demonstrated in \cref{fig:dist}.
%这里的reconstruction error是1000个全图pixel平均，单图pixel多次平均没有这么明显的差距
By contrast, epistemic uncertainty captures the model’s lack of knowledge about unfamiliar patterns, increasing significantly for samples far outside the training distribution~\cite{kendall2017uncertainties}. This property makes epistemic uncertainty more reliable for identifying diffusion-generated images. This is also validated in \cref{fig:eu}, where epistemic uncertainty more accurately distinguishes real samples from fake ones.

%当我们从x0重建一个x0'时，我需要利用公式（1)或（2）degenerate得到一个xt，然后根据公式（3）或（4）denoise。
%前向过程是一个高斯过程，会引入随机性。
%\ref{hypediff}当我们对此前向过程进行逆向时，逆向结果自然地会含有aleatoric uncertainty和epistemic uncertainty。
%多次逆向过程的var能表达aleatoric uncertainty,我们对齐进行分析，却发现其值远大于重建误差。如图（解释）
%重建误差（的采样）自身包含了不确定性
%对应修改intro
%(1)陈述diffusion reconstruction存在2中uncertain
%(2)参考hyperidfu，解释2中uncertain
%(3)引出a uncertainty不能区分ood，反而会造成混淆；e uncertainty能区分ood
%(4)给出图例证明

\subsection{Laplace Approximation for Bayesian Diffusion Models}
\label{pre-la}
To capture epistemic uncertainty, we turn the deterministic neural network $\boldsymbol{\epsilon}_\theta(\mathbf{x}_t,t)$ into a Bayesian Neural Network~(BNN) by assuming an isotropic Gaussian prior $p(\theta)$.
Let $\mathcal{D}$ be the training dataset, the target of Bayesian inference is to infer the posterior $p(\theta|\mathcal{D})$
and predict distribution of noise for noise-corrupted data $\mathbf{x}_t^\ast$ with 
\begin{equation}
 p(\boldsymbol{\epsilon}_t^\ast|\mathbf{x}_t^\ast,\mathcal{D})=\int{p(\boldsymbol{\epsilon}|\boldsymbol{\epsilon}_\theta(\mathbf{x}_t^\ast,t))p(\theta|\mathcal{D})d\theta}\,.
\end{equation}
%\epsilon也是满足高斯分布的概率模型，其实是回归任务的输出
% \begin{equation}
%  p(\boldsymbol{\epsilon|\mathbf{x}_t^\ast},\mathcal{D})=\mathbb{E}_{p(\theta|\mathcal{D})}\boldsymbol{\epsilon}_\theta(\mathbf{x}_t^\ast,t)\,.
% \end{equation}

% \begin{lemma}
%     The variance of samples from the posterior distribution $p(\theta|\mathcal{D})$ captures epistemic uncertainty. $Var(\theta) = Var(\mathbb{E_{\boldsymbol{\epsilon}}(\boldsymbol{\epsilon}_\theta)})$
% \end{lemma}
%这里D=(X,\episilon)?
Since $p(\theta|\mathcal{D})$ cannot be computed directly, Laplace Approximation~(LA) approximates it with 
\begin{equation}
\label{la_infer}
    q(\theta)=\mathcal{N}(\theta;\theta_{\mathrm{MAP}},\Sigma)\,,
\end{equation}
where $\theta_{\mathrm{MAP}}$ is the maximum a posteriori~(MAP) estimate, $\theta_{\mathrm{MAP}}=\mathrm{arg}\max_\theta(\log p(\mathcal{D|\theta})+\log(p(\theta)))$, and $\Sigma=[-\bigtriangledown _\theta^2(\log p(\mathcal{D}|\theta)+\log p(\theta))|_\theta=\theta_{\mathrm{MAP}}]^{-1}$.
Two techniques have been proposed to simplify the estimation of $\Sigma$. 
The first technique is Hessian approximations with factorization, and the most lightweight case is a diagonal factorization which ignores off-diagonal elements~\cite{lecun1989optimal}.
The second technique is the subnetwork LA, and the last-layer LA~(LLLA) is its special case which only treats the parameters of the last probabilistically~\cite{daxberger2021laplace}.
\section{Method}
\label{method}
As discussed in \cref{sec:preliminaries}, the primary goal of our proposed method is to identify diffusion-generated images by exploiting epistemic uncertainty in diffusion. We first briefly explain the learning framework, and then illustrate the process of epistemic uncertainty estimation. Finally, we explain the details of asymmetric learning.

\subsection{Overview of Framework}
\begin{figure}[t]
\centering
\centerline{\includegraphics[width=\columnwidth]{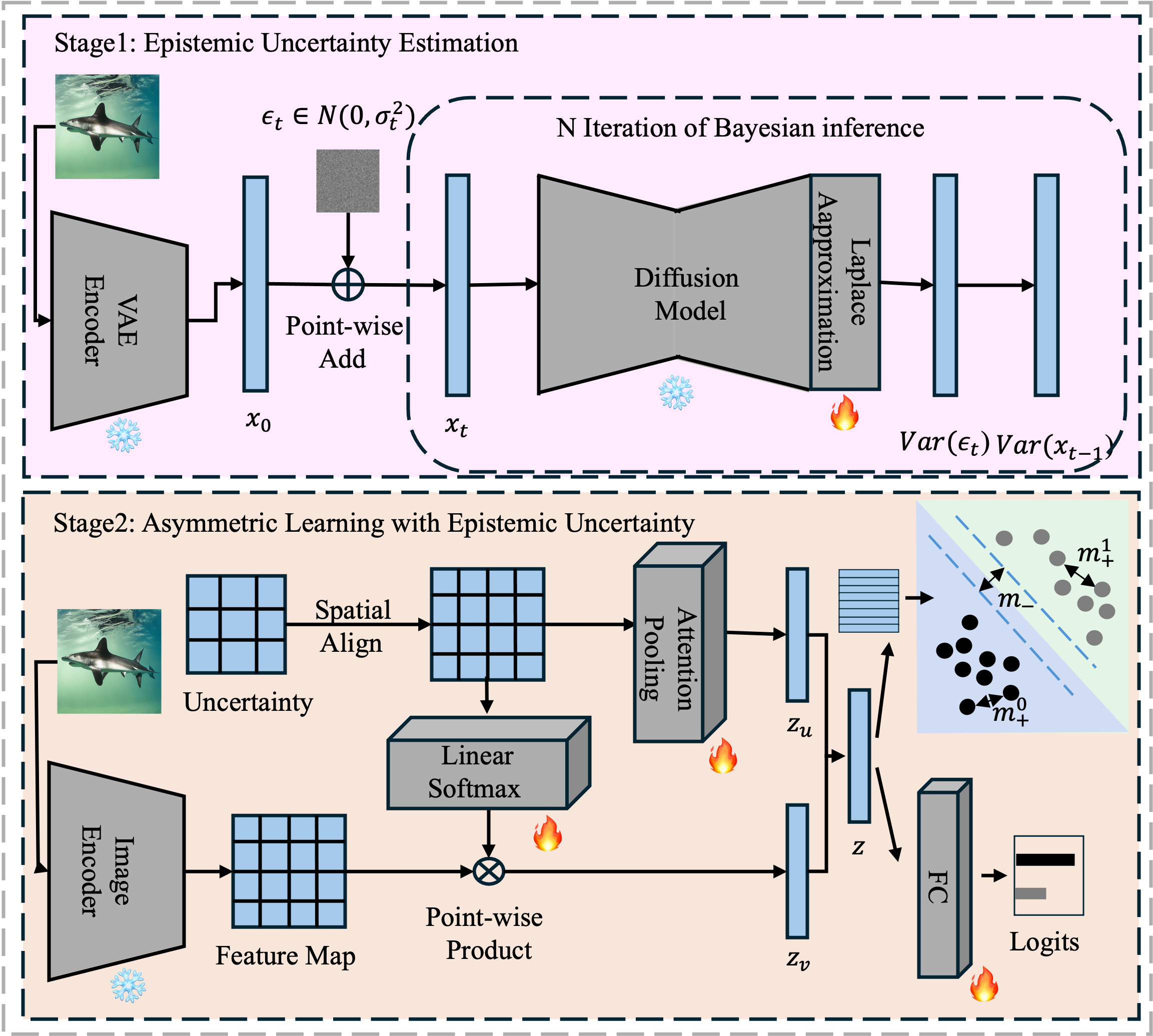}}
\insideskip
\caption{Workflow of our method. In the first stage, we utilize the Laplace approximation to estimate diffusion epistemic uncertainty. In the second stage, we exploit diffusion epistemic uncertainty to train a binary classifier with asymmetric learning.}
\label{workflow}
\outsideskip
\end{figure}

%需要给framework起个名字
\Cref{workflow} presents the workflow of our proposed method.
The framework consists of two main stages: an epistemic uncertainty estimation stage and a training stage.
%estimating diffusion epistemic uncertainty leveraging LLLA and training a classifier exploiting the estimated epistemic uncertainty.
In the uncertainty estimation stage, epistemic uncertainty is estimated in the latent space for each image using a pre-trained diffusion model.
The estimated epistemic uncertainty is then prepared for the classifier training. 
In the training stage, we incorporate the estimated uncertainty with CLIP~\cite{radford2021learning} visual feature motivated by the findings that both visual feature and diffusion measurements are important for generated image detection~\cite{luo2024lare,chendrct,ojha2023towards}.
% Our network module follows LaRE\textsuperscript{2}~\cite{luo2024lare} but in a simpler integrated manner.
%ligh-weight
%We use uncertainty in two ways: as spatial attention to extract important visual features, and as a classification feature.
Our exploitation of uncertainty is divided into two parts: as a spatial attention map to extract important visual features, and as classification features.

Let $\mathbf{u}\in \mathbb{R}^{HW\times C_1}$ be the spatially aligned epistemic uncertainty and $\mathbf{v}\in \mathbb{R}^{HW\times C_2}$ be the visual feature map.
We denote the mean epistemic uncertainty feature by $\bar{\mathbf{u}}\in \mathbb{R}^{1\times C_1}$, which is the average of $\mathbf{u}$.
We denote the global epistemic uncertainty feature by $\mathbf{z}_u$, which is simply computed using an attention pooling module on $\mathbf{u}$.
Then the refined visual feature is computed with a multi-head attention module:
\begin{equation}
\mathbf{z_v}=\text{MHA}(\bar{\mathbf{u}},\mathbf{u},\mathbf{v})\,.
\end{equation}
Finally, $\mathbf{z_u}$ and $\mathbf{z_v}$ are concatenated as the final feature.
We train a binary classifier based on this final feature.
% \begin{equation}
%     z = Concat(\mathbf{z}_u, \mathbf{z}_v)\,.
% \end{equation}

\subsection{Epistemic Uncertainty Estimation in Diffusion}
\label{sec:method-deu}
%（1）主要思路陈述。我们希望从数据的内在分布上区分fake和real。inspired by...
%（2）uncertainty估计公式推导
%（3）细节，如Hessian approximation
% Inspired by the observation that predictive ambiguity induced by aleatoric uncertainty in diffusion misleads OOD sample detection while epistemic uncertainty facilitates the detection, we propose to disentangle these two uncertainties and leverage epistemic uncertainty as a generalized representation for diffusion-generated image detection.
To estimate diffusion epistemic uncertainty, we perform the forward and reverse diffusion process in the latent space $\mathcal{X}=\{\mathbf{x}^{(i)}\}$ encoded by a pre-trained VAE~\cite{kingma2013auto} model, as stated in \cref{sec:pre-diffusion}.
\begin{lemma}
    The epistemic uncertainty in the diffusion model $\boldsymbol{\epsilon}_\theta(\mathbf{x}_t, t)$ is captured by
% \begin{equation}
% \begin{aligned}
%     Var_{\theta \sim p(\theta|\mathcal{D})}(\mathbf{x}_{t-1}|\mathbf{x},t) = \\
%     \mathbb{E}_{\boldsymbol{\epsilon}}(Var_\theta(\mathbf{x}_{t-1}|\sqrt{\alpha}_t\mathbf{x}+(1-\alpha_t)\boldsymbol{\epsilon})
% \end{aligned}   
% \end{equation}
\begin{equation}
\label{var}
\begin{aligned}
    \text{Var}_{q(\mathbf{x}_{t-1}|\mathbf{x},t)} &(\mathbf{x}_{t-1}) \\
    \propto & \text{Var}_\theta(\mathbb{E}_{\boldsymbol{\epsilon}}(\boldsymbol{\mu}_\theta(\sqrt{\alpha}_t\mathbf{x}+(1-\alpha_t)\boldsymbol{\epsilon},t)))\,.
\end{aligned}   
\end{equation}
\end{lemma}
%laplace
%uncertainty estimation
% \begin{lemma}
%     The variance of samples from the posterior distribution $p(\theta|\mathcal{D})$ captures epistemic uncertainty. $Var(\theta) = Var(\mathbb{E_{\boldsymbol{\epsilon}}(\boldsymbol{\epsilon}_\theta)})$
% \end{lemma}
% \begin{lemma}
%     For a new image $\mathbf{x}^\ast$, the distribution of the predicted noise at $t$-th step on a pre-trained Bayesian diffusion model $\boldsymbol{\epsilon}_\theta(\mathbf{x}_t, t)|_{p(\theta)\sim \mathcal{N}(0,\mathbf{I})}$ is given by
%     \begin{equation}
%     p(\boldsymbol{\epsilon}|\mathbf{x}^\ast,t)=\iint{\boldsymbol{\epsilon}_\theta(\mathbf{x}^\ast+(1-\alpha_t))\boldsymbol{\epsilon}^\ast,t)p(\theta|\mathcal{D})d\theta}d\boldsymbol{\epsilon}^\ast\,.
% \end{equation}
% \end{lemma}

% \begin{equation}
%     p(x|y) = \int{p_\theta(x|a,y)q(a|y)da}
% \end{equation}
%这里D=(X,\episilon)?
By definition, epistemic uncertainty accounts for uncertainty in the model parameters. 
As stated in \cref{pre-la}, to capture variability in the model parameters we build the Bayesian diffusion model by placing a prior distribution $p(\theta)$ over the parameters of a pre-trained diffusion model $\boldsymbol{\epsilon}_\theta(\mathbf{x}_t,t)$.
With the infered posterior $p(\theta|\mathcal{D})$, we can get the predictive distribution of the denoised image at the $(t-1)$-th step by
\begin{equation}
    \begin{aligned}
        p(\mathbf{x}_{t-1}|\mathbf{x},t)
        &=\iint p_\theta(\mathbf{x}_{t-1}|\mathbf{x}_t,\mathbf{x})p(\theta|\mathcal{D})d\theta d\boldsymbol{\epsilon}\\  
        &=\int p(\theta|\mathcal{D})\int p_\theta(\mathbf{x}_{t-1}|\mathbf{x}_t,\mathbf{x})d\boldsymbol{\epsilon}d\theta\,.
    \end{aligned}
\end{equation}
As $q(\mathbf{x}_{t-1}|\mathbf{x}_t,\mathbf{x})\sim\mathcal{N}(\boldsymbol{\tilde{\mu}}_t(\mathbf{x}_t,\mathbf{x},\tilde{\beta}_t\mathbf{I}))$~\cite{ho2020denoising}, we can get
\begin{equation}
    \begin{aligned}
        \mathbb{E}_{\boldsymbol{\epsilon}}&((\mathbf{x}_{t-1}|\mathbf{x}_t,\mathbf{x})^T (\mathbf{x}_{t-1}|\mathbf{x}_t,\mathbf{x}))\\
        & \propto
        (\mathbb{E}_{\boldsymbol{\epsilon}}(\mathbf{x}_{t-1}|\mathbf{x}_t,\mathbf{x}))^T(\mathbb{E}_{\boldsymbol{\epsilon}}(\mathbf{x}_{t-1}|\mathbf{x}_t,\mathbf{x}))\,.
    \end{aligned}
\end{equation}
%这里公式可能不能导出，\epsilon的表示含义不同
% \begin{equation}
% \begin{aligned}
%      p(\mathbf{x}_{t-1}|\mathbf{x},t)\\
%      % =\iint p(\mathbf{x}_{t-1}|\boldsymbol{\mu}_\theta(\sqrt{\alpha}_t\mathbf{x}+(1-\alpha_t)\boldsymbol{\epsilon},t))
%      % p(\theta|\mathcal{D})d\theta d\boldsymbol{\epsilon}\\  
%      =\iint\mathcal{N}(\boldsymbol{\mu}_\theta(\sqrt{\alpha}_t\mathbf{x}+(1-\alpha_t)\boldsymbol{\epsilon},t)),\sigma_t^2\mathbf{I})p(\theta|\mathcal{D})d\theta d\boldsymbol{\epsilon}\\
%      =\int p(\theta|\mathcal{D})\int \mathcal{N}(\boldsymbol{\mu}_\theta(\sqrt{\alpha}_t\mathbf{x}+(1-\alpha_t)\boldsymbol{\epsilon},t),\sigma_t^2\mathbf{I})d\boldsymbol{\epsilon}d\theta\\
% \end{aligned} 
% \end{equation}
Approximating $\boldsymbol{\tilde{\mu}}_t$ with $\boldsymbol{\mu}_\theta$ we can get the predict mean via:
\begin{equation}
\begin{aligned}
    \mathbb{E}_\theta(\mathbf{x}_{t-1})
    % = \iint (\boldsymbol{\mu}_\theta(\sqrt{\alpha}_t\mathbf{x}+(1-\alpha_t)\boldsymbol{\epsilon},t))p(\theta|\mathcal{D})d\theta d\boldsymbol{\epsilon}\\
    % =\int p(\theta|\mathcal{D})\int\boldsymbol{\mu}_\theta(\sqrt{\alpha}_t\mathbf{x}+(1-\alpha_t)\boldsymbol{\epsilon},t)d\boldsymbol{\epsilon}d\theta\\
    =\int p(\theta|\mathcal{D})\mathbb{E}_{\boldsymbol{\epsilon}}(\boldsymbol{\mu}_\theta(\sqrt{\alpha}_t\mathbf{x}+(1-\alpha_t)\boldsymbol{\epsilon},t))d\theta\,.
\end{aligned}
\end{equation}
We estimate the second raw moment in the same way:
\begin{equation}
\small
    \begin{aligned}
        &\mathbb{E}_\theta(\mathbf{x}_{t-1}^T\mathbf{x}_{t-1})\\
        =&\int p(\theta|\mathcal{D})\mathbb{E}_{\boldsymbol{\epsilon}}((\boldsymbol{\mu}_\theta(\sqrt{\alpha}_t\mathbf{x}+(1-\alpha_t)\boldsymbol{\epsilon},t))^T \\
    &\qquad\qquad\qquad\qquad(\boldsymbol{\mu}_\theta(\sqrt{\alpha}_t\mathbf{x}+(1-\alpha_t)\boldsymbol{\epsilon},t)))d\theta\\
    \propto&
    \int p(\theta|\mathcal{D})(\mathbb{E}_{\boldsymbol{\epsilon}}(\boldsymbol{\mu}_\theta(\sqrt{\alpha}_t\mathbf{x}+(1-\alpha_t)\boldsymbol{\epsilon},t))^T\\
    &\qquad\qquad\qquad\qquad(\mathbb{E}_{\boldsymbol{\epsilon}}\boldsymbol{\mu}_\theta(\sqrt{\alpha}_t\mathbf{x}+(1-\alpha_t)\boldsymbol{\epsilon},t))d\theta\,.
    \end{aligned}
\end{equation}
% \begin{equation}
%     \begin{aligned}
%         \mathbb{E}_\theta(\mathbf{x}_{t-1}^T\mathbf{x}_{t-1})=\\
%         \int p(\theta|\mathcal{D})(\mathbb{E}_{\boldsymbol{\epsilon}}(\boldsymbol{\mu}_\theta(\sqrt{\alpha}_t\mathbf{x}+(1-\alpha_t)\boldsymbol{\epsilon},t)))^T\\
%         \cdot (\mathbb{E}_{\boldsymbol{\epsilon}}(\boldsymbol{\mu}_\theta(\sqrt{\alpha}_t\mathbf{x}+(1-\alpha_t)\boldsymbol{\epsilon},t)))d\theta\\\
%         +(1-\alpha_t)???
%     \end{aligned}
% \end{equation}
% Hence, we can get the
% model’s predictive variance regarding to $\theta$ as \cref{var}.
% \begin{equation}
% \begin{aligned}
%     Var_{q(\mathbf{x}_{t-1}|\mathbf{x},t)}(\mathbf{x}_{t-1})\\
%     =
% \end{aligned}
% \end{equation}
% \begin{equation}
%     p(z)=\iint p(x,y,z)dxdy
% \end{equation}
% \begin{equation}
%     y=\int p(a)f_a(x)da
% \end{equation}
% We can get the predictive variance as ~\cref{var}.
% To estimate the
By the law of $\text{Var}(\mathbf{x})=\mathbb{E}(\mathbf{x}^T\mathbf{x})-(\mathbb{E}(\mathbf{x}))^T(\mathbb{E}(\mathbf{x}))$, we can get the predictive variance as \cref{var}.

We leverage the LLLA with diagonal factorization to approximate $p(\theta|\mathcal{D})$ as \cref{la_infer}.
The Monte Carlo method is used to approximate \cref{var}.
We sample model parameters $\theta_i\sim q(\theta)=\mathcal{N}(\theta;\theta_{\mathrm{MAP}},\Sigma)$ and noise $\boldsymbol{\epsilon}_j\sim \mathcal{N}(0,\mathbf{I})$, where $i=1,..,M$ and $j=1,...,N$.
Then epistemic uncertainty is computed as:
\begin{equation}
    U(\mathbf{x}_{t-1}|\mathbf{x},t)\\
    =\text{Var}_i(\mathbb{E}_j(\boldsymbol{\mu}_{\theta_i}(\sqrt{\alpha}_t\mathbf{x}+(1-\alpha_t)\boldsymbol{\epsilon}_j,t)))\,.
\end{equation}
We choose DDIM as the sampler and set $\sigma_t=0$ for simplicity, while other samplers, e.g., DDPM, SDE-Solver~\cite{song2020score}, DPM-Solver~\cite{lu2022dpm} is also applicable.
Accordingly, we compute $\boldsymbol{\mu}_{\theta_i}$ as \cref{xt-1}
\subsection{Asymmetric Learning}
%对于大部分结构相似的diffusion-images，train a binary classifier with celoss is enough.
%但是对于那些结构相差很大的generative model，虽然他们在数据分布上与real有差别，但是他们自身之间也存在一定的差别。heavy training with celoss leads a bias to traning data(快速过拟合)

%尽管uncertainty能表征diffusion-generated images在数据分布上的一致性，但是由于real images 分布的方差（variability?variance?），training with celoss会导致the classifier's decision boundary is unevenly bound to the fake image class. ~/cite{ojha2023towards}
%这会导致2个问题，（1）?分类器没有去学习real/fake之间数据分布的差异，而是过拟合到训练集中更容易区分real/fake的specific feature；（2）fake的特征空间被压缩的很窄，失去rundant信息支撑the distribution landscape
%这最终导致分类器的泛化性能下降，特别当生成模型结构有大的变化时
%为了解决这个问题我们使用了非对称loss,如下
%解释loss
%为了防止分类界面的倾斜，并为fake class保留足够的信息以支撑其分布空间，我们提出非对称学习框架。
%为了缓解cross-entropy的poor margin并且更准确地characterize real。
To learn an even decision boundary with larger margins while preserving the information content of the representation, we propose the Asymmetric Learning~(ASL) method.

We employ an asymmetric contrastive loss which maximizes the distance of negative pairs and minimizes the distance of positive pairs by class-specific margins.
The loss function is formulated as:
% \begin{equation}
%     \label{contrastiveloss}
%     \begin{aligned}
%         \ell_m=\frac{1}{N}\sum_{i=1}^N\mathbb{I}[y_i^{(1)}=y_i^{(2)}=c]\cdot \max(0, d_W(i)-m_+^c)\\
%         +\mathbb{I}[y_i^{(1)}\ne y_i^{(2)}]\cdot \max(0,m_--d_W(i))\,,
%     \end{aligned}
% \end{equation}
\begin{equation}
    \begin{aligned}
        \ell_m=\frac{1}{N}&\sum_{i=1}^N\mathbb{I}[y_i^{(1)}=y_i^{(2)}=c]\cdot \max(0, m^c-s_W(i))\\
        &+\mathbb{I}[y_i^{(1)}\ne y_i^{(2)}]\cdot \max(0,s_W(i))\,,
    \end{aligned}
    \label{contrastiveloss}
\end{equation}
% where $N$ is the total number of sample pairs, $y_i$ is the binary label for each sample~(0 denotes the real class), $d_W(i)$ is the Euclidean distance between the samples in each pair.
% $m_-$ is the margin for negative pairs and is set to 1.0 by default in our experiments.
% $m_+^c$ is the margin for positive pairs, and we set a larger margin for the fake class.
% That is, we separate negative pairs by a margin and compress the embedding space of the real class more strongly to alleviate the skewness of the decision boundary.
where $N$ is the total number of sample pairs, $y_i$ is the binary label for each sample~(0 denotes the real class), $s_W(i)$ is the Cosine similarity between the samples in each pair, $m^c$ is the margin specific to the class $c$.
%, and we set a larger margin for the fake class.
% That is, we the feature space of the real class is represented with a broader landscape .
Characterizing the wide range of features exhibited by the real class presents a significant challenge, making it easier for the model to focus on the simpler artifacts and resulting in a skewed decision boundary~\cite{ojha2023towards}. 
To address this issue, we propose the implementation of a smaller similarity margin specifically for the real class. 
The margin for the fake class $m^1$ is set to 1 by default, which is inherited from the general contrastive loss. 
We modify $m^0$ to enhance generalizability.
% In addition, we add a regularization term to further maximize the information content of the embedding for the fake class, which is borrowed from VICReg~\cite{bardes2021vicreg}.
% We keep only the covariance term:
% \begin{equation}
%     \label{reg}
%     \begin{aligned}
%         \ell_r=\frac{1}{d}\sum_{i\ne j}(\frac{1}{n-1}\sum_{i=1}^N(z_i-\bar{z})(z_i-\bar{z})^T)\,,
%     \end{aligned}
% \end{equation}
% where $z_i$ is the embedding of any sample from the fake class, and $\bar{z}$ is the mean embedding within a batch.

The overall loss function is a weighted average of cross-entropy loss, asymmetric contrastive loss:
%and the regularization term:
\begin{equation}
\label{loss_total}
\ell(W)=\ell_c(W)+\lambda\ell_m(W)\,,
\end{equation}
where $\ell_c$ is the cross-entropy loss, $\lambda$ is the hyper-parameter controlling the importance of $\ell_m$.
% \begin{equation}
% \label{loss_total}
% \ell(W)=\ell_c(W)+\lambda_m\ell_m(W)+\lambda_r\ell_r(W)\,,
% \end{equation}
% where $\ell_c$ is the cross-entropy loss, $\lambda_m$ and $\lambda_r$ are hyper-parameters controlling the importance of each term.

%我们这里使用非对称metri loss，使类间的距离最大，real类中距离最小，而fake中的类中距离保持在一个margin.
%同时我们使用xxx正则使fake空间保持足够的信息
%实验表明，class-weighted celoss主要是提高ACC，而lm和lr则可以提高AUC
%\subsection{Implementation Details}
\section{Experiments}
\subsection{Datasets and Evaluation Metrics}
Following LaRE\textsuperscript{2}~\cite{luo2024lare} and DRCT~\cite{chendrct}, we performed evaluations on two large-scale datasets: GenImage~\cite{Zhu2023GenImageAM} and DRCT-2M~\cite{chendrct}.
GenImage comprises 2,681,167 images, segregated into 1,331,167 real and 1,350,000 fake images.
GenImage employed all real images from ImageNet~\cite{deng2009imagenet}.
All fake images were generated following the template prompt ``photo of [class]'', where ``[class]'' was substituted by one of 1000 distinct image classes provided in ImageNet.
Eight generative models were used for image generation, namely BigGAN~\cite{Brock2018LargeSG}, GLIDE~\cite{Nichol2021GLIDETP}, VQDM~\cite{Gu_Chen_Bao_Wen_Zhang_Chen_Yuan_Guo_2022}, Stable Diffusion V1.4\&V1.5~\cite{Rombach_Blattmann_Lorenz_Esser_Ommer_2022}, ADM~\cite{Dhariwal_Nichol_2021}, Midjourney~\cite{midjourney}, and Wukong~\cite{Wukong}.
DRCT-2M consists of two parts: 1,920,000 images generated by various diffusion-based generative models and the real images from MSCOCO~\cite{Lin2014MicrosoftCC}.
% To generate images, DRCT-2M employs 10 types of the Stable Diffusion models~\cite{Rombach_Blattmann_Lorenz_Esser_Ommer_2022} using a text-to-image process and 3 types ControlNet~\cite{Zhang2023AddingCC}  using an image-to-image process.
DRCT-2M generated images using two kinds of processes: a text-to-image process covering 10 types of the Stable Diffusion models~\cite{Rombach_Blattmann_Lorenz_Esser_Ommer_2022} with prompts derived from MSCOCO and an image-to-image process covering 3 types of Stable Diffusion models and 3 types of ControlNet~\cite{Zhang2023AddingCC}.
We adopted the official division of these two datasets.
Following previous works~\cite{wang2023dire}, we used accuracy~(ACC) with a threshold of 0.5 and average precision~(AP) as the metric to evaluate detection performance in our experiments.
%Raising the Bar里有提到acc threshold随着数据增加的影响

\subsection{Implementation Details}
The proposed method consists of the estimation of DEU and the training of a classifier.
To estimate DEU, we used Stable Diffusion V1.5 with a step size of $t=200$ by default.
The prompt was ``a photo'' for all images.
During training and testing phases, the classifier took input images of size 224 × 224.
CLIP pre-trained ResNet50~\cite{he2016deep} was utilized to extract image features.
The batch size was set to 48 and the learning rate was set to $1\text{e}{-4}$.
We selected $\lambda$ based on the SDv1.5 subset of GenImage, and set it to 0.5.
We found it works well across datasets.
% For GenImage, we train eight models on eight subsets respectively.
% For DRCT-2M, 
All experiments were conducted on NVIDIA Tesla V100 GPUs.

\subsection{Comparison with the State of the Arts}

%补充画图和解释说明
%是否需要解释为什么现有方法的AP值很高，但是ACC低？
%LaRE已经取得了near-perfect AP,但是其最优阈值在不同subset上variable，我们推测其可能得原因是
%解释为什么不跟DRCT比较？DRCT训练的时候会使用SD DR的图片（supp 里面补充在sd1.4上训练的实验？）
%BigGAN平均测试并不差？
%结论：
%（1）对于在diffusion generated subsets（除了MJ）上训练，ALDEU能稳定区分真伪
%（2）
\begin{table*}[t]
\centering
\resizebox{0.85\textwidth}{!}{
\begin{tabular}{clccccccccc}
\toprule
DRCT DR &Method & Midjourney & SDV1.4 & SDV1.5 & ADM & GLIDE & Wukong & VQDM & BigGAN & Avg. \\
\midrule
\multirow{6}{*}{w/o} & F3Net~{\tiny [ECCV 2020]}   & 55.1 & 73.1 & 73.1 & 66.5 & 57.8 & 72.3 & 62.1 & 56.5 & 64.6 \\
& GramNet~{\tiny [CVPR 2020]} & 58.1 & 72.8 & 72.7 & 58.7 & 65.3 & 71.3 & 57.8 & 61.2 & 64.7 \\
& UnivFD~{\tiny [CVPR 2023]}  &\underline{70.1}&74.8&75.0&62.9&77.6&72.2&64.8&60.4&69.7 \\
& DIRE~{\tiny [ICCV 2023]}    & 65.0 & 73.7 & 73.7 & 61.9 & 69.1 & 74.3 & 63.4 & 56.7 & 67.2 \\
& LaRE\textsuperscript{2}~{\tiny [CVPR 2024]} & 66.4 & \underline{87.3} & \underline{87.1} & \underline{66.7} & \underline{81.3} & \underline{85.5} & \underline{84.4} & \underline{74.0} & \underline{79.1} \\
\cmidrule(lr){2-11}
& Ours    & \textbf{80.7} & \textbf{89.1} & \textbf{88.6} & \textbf{78.9} & \textbf{88.4} & \textbf{88.1} & \textbf{86.6} & \textbf{83.7} & \textbf{85.6} \\
\midrule
\multirow{3}{*}{w/}  & DRCT/ConvB~{\tiny [ICML 2024]} & 78.2 & \textbf{97.6} & \textbf{97.1} & 74.2 & 75.3 & \underline{96.0} & 72.3 & 67.6 & 82.3 \\
& DRCT/UniFD~{\tiny [ICML 2024]} & \underline{83.8} & 93.1 & 92.6 & \underline{83.2} & \underline{89.5} & 92.9 & \underline{91.8} & \textbf{86.0} & \underline{89.1}\\
\cmidrule(lr){2-11}
& Ours & \textbf{86.4} & \underline{96.5} & \underline{96.2} & \textbf{85.3} & \textbf{94.4} & \textbf{96.2} & \textbf{93.1} & \underline{84.2} & \textbf{91.5} \\
\bottomrule
\end{tabular}
}
\insideskip
\caption{Accuracy~(ACC, \%) comparisons on GenImage test subsets. Eight models are trained on eight generators respectively. All the eight models are tested on the specified test subsets, and averaging the accuracy scores yields the final results.}
\label{com-genimage}
\outsideskip
\end{table*}

\begin{table*}[t]
\centering
\resizebox{\textwidth}{!}{%
\begin{tabular}{clcccccccccccccccccc}
\toprule
% Method & LDM & SDv1.4 & SDv1.5 & SDv2 & SDXL & SDXL- Refiner & SD-Turbo & SDXL-Turbo & LCM-SDv1.5 & LCM-SDXL & SDv1-Ctrl & SDv2-Ctrl & SDXL-Ctrl & SDv1-DR & SDv2-DR & SDXL-DR & Avg. \\
\multirow{3}{*}{DRCT DR}
&\multicolumn{1}{l}{\multirow{3}{*}{Method}} 
&\multicolumn{1}{c}{\multirow{3}{*}{DR}} 
&\multicolumn{6}{c}{SD Variants}
&\multicolumn{2}{c}{Turbo Variants}&\multicolumn{2}{c}{LCM Variants}&\multicolumn{3}{c}{ControlNet Variants}&\multicolumn{3}{c}{DR Variants}&\multicolumn{1}{c}{\multirow{3}{*}{Avg.}}\\
\cmidrule(lr){4-9} \cmidrule(lr){10-11} \cmidrule(lr){12-13} \cmidrule(lr){14-16} \cmidrule(lr){17-19}
&&& \multicolumn{1}{c}{LDM} & \multicolumn{1}{c}{SDv1.4} & \multicolumn{1}{c}{SDv1.5} & \multicolumn{1}{c}{SDv2} & \multicolumn{1}{c}{SDXL} & \multicolumn{1}{c}{\makecell[c]{SDXL- \\ Refiner}} & \multicolumn{1}{c}{\makecell[c]{SD- \\ Turbo}} & \multicolumn{1}{c}{\makecell[c]{SDXL- \\ Turbo}} & \multicolumn{1}{c}{\makecell[c]{LCM- \\ SDv1.5}} & \multicolumn{1}{c}{\makecell[c]{LCM- \\ SDXL}} & \multicolumn{1}{c}{\makecell[c]{SDv1- \\ Ctrl}} & \multicolumn{1}{c}{\makecell[c]{SDv2- \\ Ctrl}} & \multicolumn{1}{c}{\makecell[c]{SDXL- \\ Ctrl}} & \multicolumn{1}{c}{\makecell[c]{SDv1- \\DR}} & \multicolumn{1}{c}{\makecell[c]{SDv2- \\ DR}} & \multicolumn{1}{c}{\makecell[c]{SDXL- \\ DR}}  \\
\midrule
\multirow{7}{*}{w/o} & F3Net  &-& \textbf{99.9} & 99.8 & \underline{99.8} & 88.7 & 55.9 & 87.4 & 68.3 & 63.7 & 97.7 & 55.0 & 98.0 & 72.4 & 82.0 & \textbf{65.4} & 50.4 & 50.3 & 77.1\\
& GramNet &-& \underline{99.4} & 99.0 & 98.8 & 95.3 & 62.6 & 80.7 & 71.2 & 69.3 & 93.1 & 57.0 & 90.0 & 75.6 & 82.7 & 51.2 & 50.0 & 50.1 & 76.6\\  
& UnivFD &-& 98.3 & 96.2 & 96.3 & 93.8 & 91.0 & 93.9 & 86.4 & 85.9 & 90.4 & 89.0 & 90.4 & 81.1 & 89.1 & 52.0 & 51.0 & 50.5 & 83.5\\
& DIRE &SDv1& 98.2 & \underline{99.9} & \textbf{100.0} & 68.2 & 53.8 & 71.9 & 58.9 & 54.4 & \textbf{99.8} & 59.7 & \textbf{99.7} & 64.2 & 59.1 & 52.0 & 50.0 & 50.0 & 71.2\\ 
% DIRE & SDv2 & 54.6 & 75.8 & 76.0 & 99.8 & 59.9 & 93.0 & 99.7 & 57.5 & 87.2 & 72.5 & 67.8 & 99.6 & 64.4 & 49.9 & 52.4 & 49.9 & 72.5 & 77.5 \\
& LaRE\textsuperscript{2} & SDv1 & \underline{99.4} & \textbf{100.0} & \textbf{100.0} & 96.3 & 97.2 & 97.6 & 98.6 & 86.4 & 96.1 & \underline{94.2} & 96.4 & \underline{99.2} & 96.2  & 49.5 & 50.6 & 50.0 & \underline{88.0}\\
\cmidrule(lr){2-20}
& Ours & SDv1 &99.2 & 99.2 & 99.2 & \underline{99.2} & \underline{99.2} &	\underline{99.1} & \underline{99.2} & \underline{99.1} & \underline{99.2} & \textbf{99.2} & \underline{99.2} & \underline{99.2} & \underline{99.2} & \underline{54.7} & \textbf{53.1} & \underline{51.1} & \textbf{90.5}\\
& Ours & SDv2 &99.2&99.1& 99.2&\textbf{99.4}&\textbf{99.4}&\textbf{99.2}&\textbf{99.3}&\textbf{99.2}&99.1&\textbf{99.2}&\underline{99.2}&\textbf{99.4}&\textbf{99.4}&54.1&\underline{51.2}&\textbf{52.2}&\textbf{90.5} \\
\midrule
\multirow{6}{*}{w/} & DRCT/Conv-B & SDv1 & \textbf{99.9} & \textbf{99.9} & \textbf{99.9} &	96.3 &	83.9 & 85.6 & 91.9 & 70.0 & \textbf{99.7} & 78.8 & \textbf{99.9} & 95.0 & 81.2 & \textbf{99.9} & \underline{95.4} & 75.4 & 90.8 \\
& DRCT/Conv-B &SDv2& \underline{99.7} & 98.6 & 98.5 & 99.9 & 96.1 & 98.7 & \textbf{99.6} & 83.3 & 98.5 & 93.8 & 96.7 & \textbf{99.9} & 97.7 & 93.9 & \textbf{99.9} & \textbf{90.4} & 96.6 \\
& DRCT/UniFD & SDv1 & 96.7 & 96.3 & 96.3 & 94.9 & 96.2 & 93.5 & 93.4 & 92.9 & 91.2 & 95.0 & 95.6 & 92.7 & 92.0 & 94.1 & 69.6 & 57.4 & 90.5 \\
& DRCT/UniFD &SDv2& 94.5 & 94.4 & 94.2 & 95.1 & 95.6 & 95.4 & 94.8 & 94.5 & 91.7 & 95.5 & 93.9 & 93.5 & 93.5 & 84.3 & 83.2 & 67.6 & 91.4 \\
\cmidrule(lr){2-20}
& Ours & SDv1 & \textbf{99.9} & \textbf{99.9} & \textbf{99.9} & \underline{99.9} & \underline{99.9} & \underline{99.4} & 99.4 & \underline{99.3} & 99.1 & \textbf{99.4} & \textbf{99.9} & \underline{99.4} & \underline{99.4} & \textbf{99.9} & 94.2 & \underline{90.}1 & \underline{98.7} \\
& Ours & SDv2 & 99.4 & \underline{99.7} & \underline{99.8} & \textbf{100.0} & \textbf{100.0} & \textbf{99.5} & \underline{99.5} & \textbf{99.5} & \underline{99.3} & \underline{99.2} & \underline{99.4} & \textbf{99.9} & \textbf{99.9} & \underline{99.2} & 95.3 & \underline{90.1} & \textbf{98.8} \\
\bottomrule
\end{tabular}%
}
\insideskip
\caption{Accuracy~(ACC, \%) comparisons on DRCT-2M. All methods are only trained on SDv1.4 and evaluated on different test subsets on DRCT-2M. For DRCT and our method, we report performance metrics utilizing different diffusion reconstruction model~(DR), specifically SDv1 and SDv2.}
\label{com-drct}
\outsideskip
\end{table*}

%补充对于Ours with DRCT DR的说明
We compared our method with several state-of-the-art image generation detection approaches: F3Net~\cite{qian2020thinking}, GramNet~\cite{liu2020global}, UnivFD~\cite{ojha2023towards}, DIRE~\cite{wang2023dire}, LaRE\textsuperscript{2}~\cite{luo2024lare}, and DRCT~\cite{chendrct}.
All experimental setups followed the guidelines established by the GenImage~\cite{Zhu2023GenImageAM} and DRCT-2M~\cite{chendrct} benchmarks.
For the GenImage dataset, we trained a classification model on each of the eight subsets, each corresponding to a different generation method. 
Each trained model was then evaluated on all eight test subsets.
For the DRCT-2M dataset, we trained a classification model on the SDv1.4 subset of DRCT-2M and test it on all the test subsets.
A notable distinction in the DRCT's setup is the inclusion of Stable Diffusion inpainting images (referred to as DRCT DR) as hard samples in the training set. 
To ensure a fair comparison, we aligned the settings and introduced an additional experiment that incorporated images reconstructed using the DRCT inpainting process for a direct comparison with DRCT.

The averaged results over eight trained models on each test subsets of the GenImage dataset are presented in \cref{com-genimage}.
In the scenario where DRCT DR was not applied, our method outperformed the state-of-the-art by 6.5\% in average accuracy~(ACC). When integrated with DRCT DR, our method shown an additional gain, outperforming DRCT by 2.4\% in average ACC. 
% \blue{The improvements observed with DRCT DR are particularly pronounced when training on subsets that include generators significantly different from Stable Diffusion.}
Similar results are observed on the DRCT-2M dataset, as shown in \cref{com-drct}. Without DRCT DR, our method exceeded the state-of-the-art by achieving a 2.5\% increase in average ACC. With DRCT DR, our method surpassed DRCT by 2.2\% in average ACC. The improvements from incorporating DRCT DR were most notable in the DR variant subsets, which were reconstructed from real images using the DRCT inpainting process.

% \blue{In the cross-dataset experiment, as illustrated in \cref{com-cross}, our method achieves superior performance with an accuracy of \red{89.4\%} and an average precision of \red{94.2\%}, demonstrating its enhanced generalizability compared to other methods.} \red{Add description about DRCT DR.}

%大部分已有方法在seen generators均有好的效果，我们的提升主要来源于对于unseen的泛化性。
%补充cross dataset

\subsection{Generalizability Across Generators}

\begin{figure*}[t]
  \centering
  \begin{subfigure}{0.49\linewidth}
    \includegraphics[width=1.0\columnwidth]{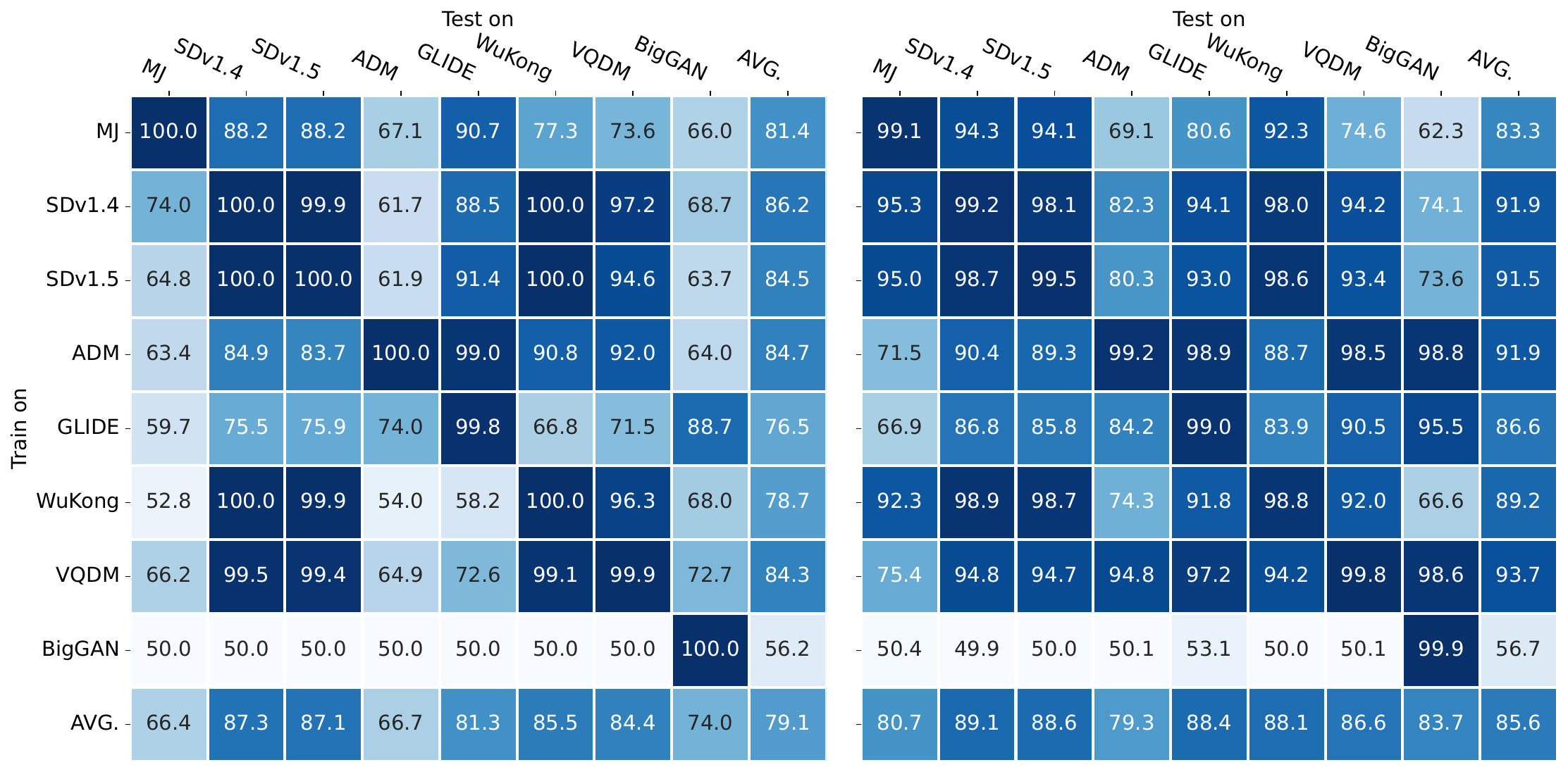}   
    \caption{Comparison on ACC. Left: baseline, right: ours.}
    \label{fig:com_acc}
  \end{subfigure}
  \hfill
  \begin{subfigure}{0.49\linewidth}
\includegraphics[width=1.0\columnwidth]{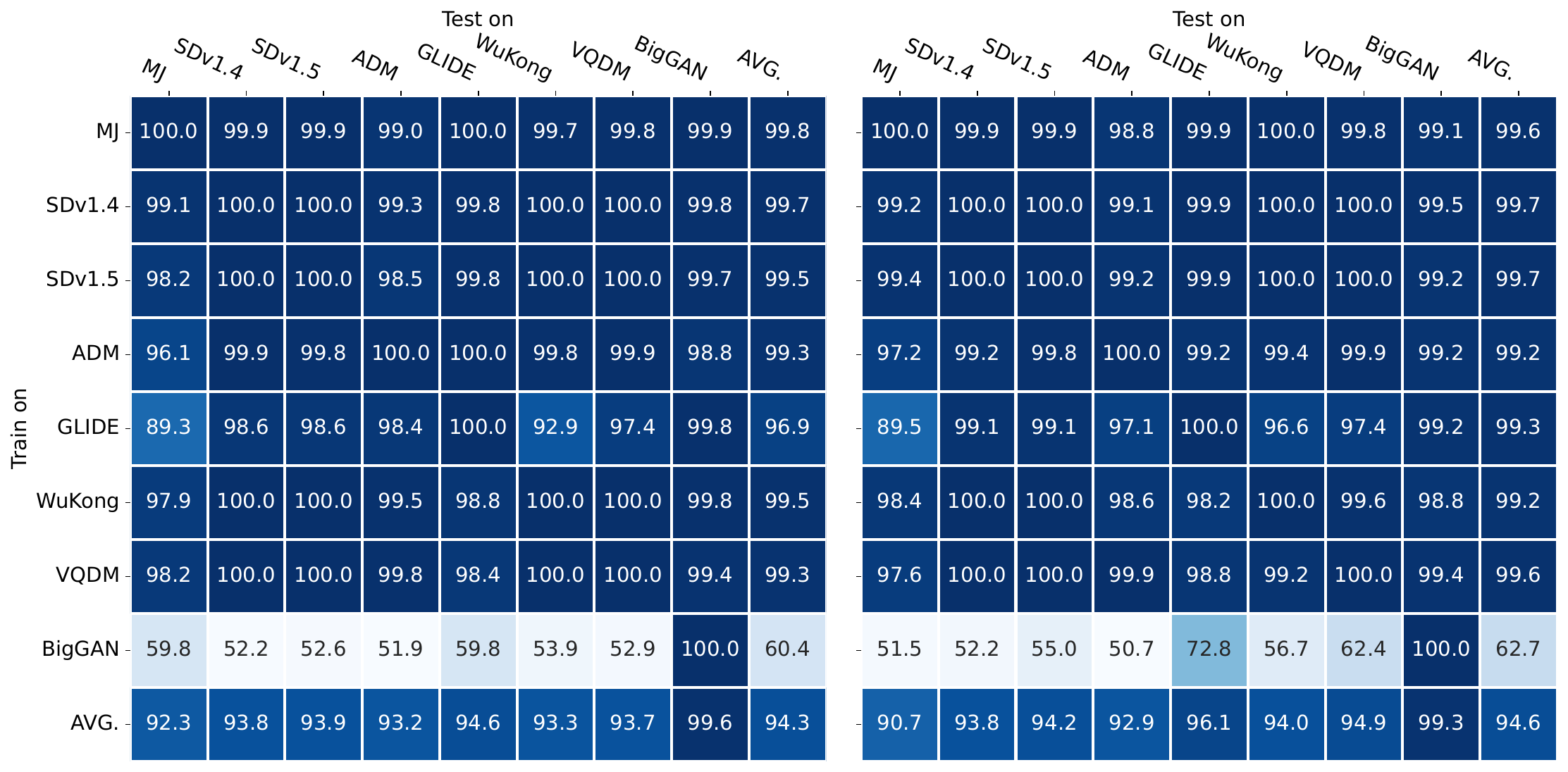}
    \caption{Comparison on AP. Left: baseline, right: ours.}
    \label{fig:com_ap}
  \end{subfigure}
\insideskip
  \caption{Results of cross-validation on GenImage. We train eight models on eight subsets respectively, each corresponding to a different generator. For both LaRE\textsuperscript{2} and our method, accuracy~(ACC, \%) and average precision~(AP, \%) are reported.}
  \label{fig:comp}
\outsideskip
\end{figure*}

We conducted generalizability comparisons across generators on GenImage and DRCT-2M, respectively.

\textbf{Comparison on GenImage}
The results of cross-validation on different training and testing subsets are detailed in \cref{fig:comp}.
Both LaRE\textsuperscript{2} and our method achieved high accuracy with images that shared the same generation method as the training set. 
This indicates that detecting generated images from seen generators is relatively straightforward.
However, generalizing the detector to unseen generators becomes progressively more challenging as the differences between the generators increase. Compared to LaRE\textsuperscript{2}, our method experienced less performance degradation across subsets with different generators, achieving an overall average ACC improvement from 79.1\% to 85.6\%.
Notably, when trained on the BigGAN subset, both methods performed poorly on diffusion-generated images. This suggests a significant distribution gap between images generated by GANs and those produced by diffusion models.
% \blue{which is ultimately reflected in the epistemic uncertainty of the diffusion-generated images.}

\textbf{Comparison on DRCT-2M}
\Cref{com-drct} reports performance comparisons across different test subsets of the DRCT-2M dataset. Consistent with prior observations on GenImage, existing methods achieve excellent performance on seen generators. For unseen generators, methods leveraging diffusion reconstruction error~(LaRE\textsuperscript{2}) and those incorporating diffusion-reconstructed images as additional training data~(DRCT) show improved performance, particularly when the diffusion models closely resemble those encountered during training.
However, a more prevalent and challenging real-world scenario involves detecting images generated by substantially altered and previously unseen diffusion models, such as SDXL, SDXL-Turbo, and LCM-SDXL. Under these conditions, all methods typically exhibit notable performance degradation.
In contrast, our proposed method consistently maintains near-perfect detection accuracy across most diffusion-generated image types, except for the DR variants. This robust generalization capability results in an overall improvement in average accuracy from 88.0\% to 90.5\%.

Like other state-of-the-art methods, our approach initially struggled with DR variants. However, by incorporating DRCT training samples, specifically the DRCT DR images, our method not only successfully overcomes this limitation but also surpasses DRCT by an additional 2.2\% in accuracy. This strategic adjustment highlights our method’s adaptability and effectiveness in handling diverse generation scenarios.
Moreover, DRCT’s performance significantly depends on the choice of reconstruction models, \ie, SDv1 or SDv2, whereas our method demonstrates substantially improved robustness.

\subsection{Generalizability Across Datasets}

\begin{table*}[t]
\centering
\resizebox{0.95\textwidth}{!}{%
\begin{tabular}{clccccccccc}
\toprule
DRCT DR & Method & Midjourney & SDV1.4 & SDV1.5 & ADM & GLIDE & Wukong & VQDM & BigGAN & Avg. \\
\midrule
% Conv-B & 68.9/78.4 & 96.2/98.2 & 96.1/98.2 & 52.7/60.4 & 52.4/62.4 & 85.8/92.4 & 53.2/59.8 & 50.0/51.2 & 69.4/75.1 \\
\multirow{6}{*}{w/o}
& F3Net~{\tiny [ECCV 2020]} & 71.7/80.4 & \underline{97.5}/\underline{99.0} & \underline{96.7}/\underline{98.8} & 55.9/66.1 & 62.2/74.2 & \underline{88.1}/\underline{93.2} & \underline{62.8}/\underline{73.1} & 50.1/50.5 & \underline{73.2}/\underline{79.4} \\
& GramNet~{\tiny [CVPR 2020]} & 70.2/\underline{81.3} & 86.5/92.2 & 86.1/91.6 & 52.2/62.5 & 53.5/65.4 & 76.3/88.7 & 54.2/64.6 & 49.8/50.7 & 66.1/74.6 \\
& UniFD~{\tiny [CVPR 2023]} & \underline{73.6}/80.1 & 74.2/84.2 & 74.2/83.9 & \underline{56.3}/64.2 & \underline{70.8}/\underline{80.2} & 73.3/82.5 & 56.9/64.8 & \underline{61.2}/\underline{72.4} & 67.6/76.5   \\
& DIRE~{\tiny [ICCV 2023]} & 52.4/54.6 & 56.1/60.2 & 55.7/60.0 & 50.2/53.2 & 50.4/56.2 & 54.2/60.2 & 49.2/55.4 & 49.2/52.4 & 52.2/56.5  \\
& LaRE\textsuperscript{2}~{\tiny [CVPR 2024]} & 56.2/71.0 & 55.1/61.6 & 54.5/61.5 & 51.3/\underline{65.6} & 60.4/75.4 & 53.3/62.0 & 52.8/65.9 & 46.1/57.2 & 53.7/65.0 \\
\cmidrule(lr){2-11}
& Ours   & \textbf{92.2}/\textbf{94.4} & \textbf{97.6}/\textbf{99.1} & \textbf{97.2}/\textbf{99.1} & \textbf{79.4}/\textbf{90.6} & \textbf{90.1}/\textbf{94.2} & \textbf{96.8}/\textbf{98.4} & \textbf{91.4}/\textbf{94.6} & \textbf{71.6}/\textbf{86.1} & \textbf{89.5}/\textbf{94.5} \\ %& \textbf{93.2} & \textbf{89.0} & \textbf{88.6} & \textbf{78.9} & \textbf{88.4} & \textbf{88.0} & \textbf{86.5} & \textbf{83.6} & \textbf{89.4} \\
\midrule
\multirow{3}{*}{w/} & DRCT/Conv-B~{\tiny [ICML 2024]} & \textbf{94.6}/\textbf{98.2} & \textbf{99.6}/\textbf{99.9} & \textbf{99.4}/\textbf{99.9} & 65.8/78.2 & 73.2/88.4 & \textbf{99.4}/\textbf{99.9} & 77.8/89.6 & 60.4/76.5 & 83.8/91.3 \\
& DRCT/UniFD~{\tiny [ICML 2024]} & 86.1/93.2 & 93.4/97.4 & 93.2/97.1 & \underline{74.2}/\underline{82.3} & \underline{85.1}/\underline{90.1} & 93.2/96.8 & \underline{89.6}/\underline{94.2} & \textbf{86.2}/\textbf{91.8} & \underline{87.6}/\underline{92.9} \\
\cmidrule(lr){2-11}
& Ours  &\underline{92.7}/\underline{95.1} & \underline{98.1}/\underline{99.2} & \underline{97.5}/\underline{99.2} & \textbf{78.8}/\textbf{90.2} & \textbf{89.7}/\textbf{94.2} & \underline{97.2}/\underline{99.2} & \textbf{90.7}/\textbf{94.4} & \underline{75.8}/\underline{90.5} & \textbf{90.1}/\textbf{95.3}\\
\bottomrule
\end{tabular}
}
\insideskip
\caption{Accuracy~(ACC, \%) / average precision~(AP, \%) comparisons of generalizability across datasets. All methods are trained on DRCT-2M/SDv1.4 using SDv1 as the diffusion reconstruction model and evaluated on different testing subsets of GenImage.}
\label{com-cross}
\outsideskip
\end{table*}

To further evaluate the generalizability of our proposed method, we conducted cross-dataset experiments following DRCT.  Specifically, we trained a classification model on the SDv1.4 subset of DRCT-2M and evaluated it across all subsets of GenImage. The cross-dataset results are summarized in~\cref{com-cross}, showing noticeable performance degradation for existing methods.
For example, LaRE\textsuperscript{2} achieved 86.2\% ACC when trained on the SDv1.4 subset of GenImage~(as presented in \cref{fig:comp}), but its performance dropped significantly to 53.7\% ACC on DRCT-2M (\cref{com-cross}), representing a 32.5\% reduction. Similarly, a 34.7\% drop in AP (from 99.7\% to 65.0\%) was observed. Overall, methods based on diffusion reconstruction errors, such as DIRE and LaRE\textsuperscript{2}, encountered substantial challenges in maintaining performance across different datasets.

In contrast, our method demonstrated stronger generalization, with smaller declines in both ACC and AP metrics under identical cross-dataset conditions (\cref{com-cross}, \cref{fig:comp}). These results underline the advantage of leveraging diffusion epistemic uncertainty, which offers improved robustness across varying image content.
Additionally, despite its diffusion focus and orthogonality to unified schemes, DEUA stays competitive and is an effective module in larger frameworks (\cref{app:gen}).

% The results of the cross-dataset evaluation are presented in \cref{com-cross}. 
% With the exception of DRCT and our method, existing detectors exhibit significant performance degradation across all subsets. 
% Notably, DIRE and LaRE\textsuperscript{2}, which rely on diffusion reconstruction error for detecting generated images, recorded the lowest detection ACC, both falling below 60\%.
% Specifically, LaRE\textsuperscript{2} experienced a 32.5\% decrease in ACC and a 34.7\% decrease in AP compared to results trained on the SDv1.4 subset of GenImage, as shown in \cref{fig:comp}.
% This decline in performance was evident even when identifying images from seen generators, suggesting a potential correlation between diffusion reconstruction error and the specific content of the images.

% In contrast, our method showed only a marginal decrease of 2.4\% in ACC and 5.2\% in AP compared to results trained on the same SDv1.4 subset of GenImage, as shown in \cref{fig:comp}.
% This indicates that diffusion epistemic uncertainty is less influenced by the image content, thereby improving its generalizability across diverse scenarios.

\subsection{Ablation Study}

\begin{table*}[t]
\centering
\resizebox{0.95\textwidth}{!}{%
\begin{tabular}{cccccccccccc}
\toprule
Method & DEU & ASL & Midjourney & SDV1.4 & SDV1.5 & ADM & GLIDE & Wukong & VQDM & BigGAN & Avg. \\
\midrule
A & & &62.8/82.1&99.9/100.0&99.9/100.0&57.2/88.3&78.4/94.5&99.3/99.9&60.2/88.2&50.8/72.6&76.1/90.7\\
B & $\surd$ & &94.1/99.2&99.6/100.0&99.8/100.0&78.5/99.2&91.8/99.6&99.2/99.9&92.8/99.9&66.4/92.1&90.3/98.7  \\
C & & $\surd$ & 71.5/86.2 & 98.4/99.8&99.2/100.0&64.6/91.5&80.2/94.8&98.2/99.6&68.6/88.2&72.8/90.2&81.7/93.8\\
D & $\surd$ & $\surd$&95.0/99.4&98.7/100.0&99.5/100.0&80.3/99.2&93.0/99.9&98.6/100.0&93.4/100.0&73.6/99.2&91.5/99.7 \\
\bottomrule
\end{tabular}
}
\insideskip
\caption{Ablative study results on GenImage test subsets.}
\label{com-ablation}
\outsideskip
\end{table*}

\begin{figure}[t]
  \centering
  \begin{subfigure}{0.49\linewidth}
    \includegraphics[width=1.0\columnwidth]{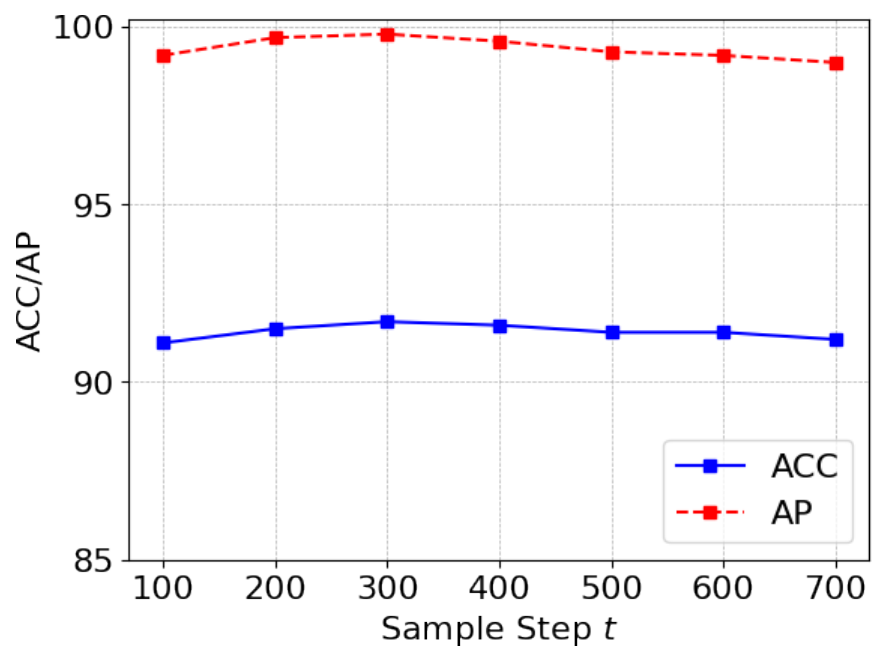}   
    \caption{Accuracy~(ACC, \%) / average precision~(AP, \%) with reconstruction step $t$ range from 100 to 700.}
    \label{fig:step}
  \end{subfigure}
  \hfill
  \begin{subfigure}{0.49\linewidth}
  \includegraphics[width=1.0\columnwidth]{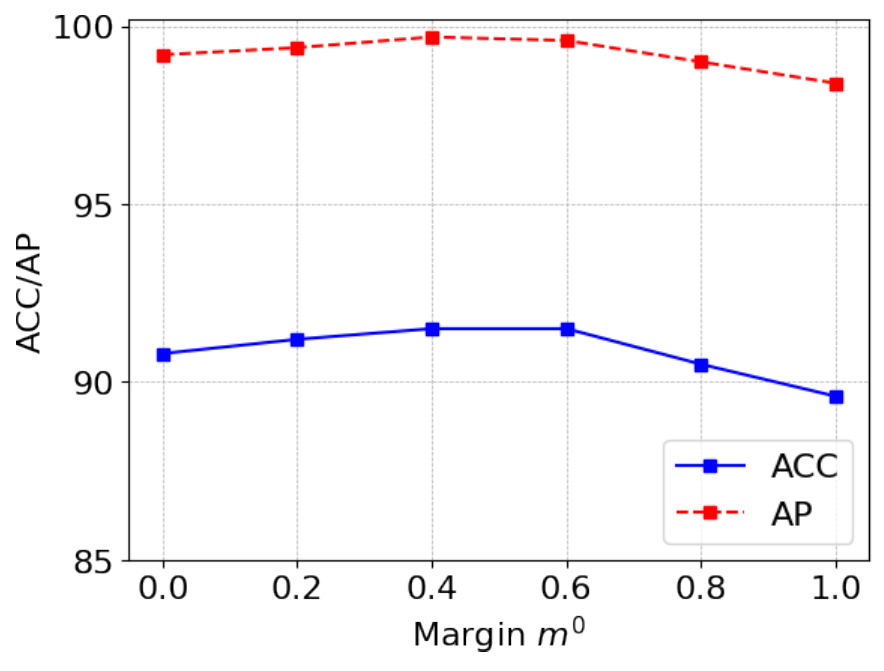}
    \caption{Accuracy~(ACC, \%) / average precision~(AP, \%) with the real-class margin $m^0$ range from 0.0 to 1.0.}
    \label{fig:margin}
  \end{subfigure}
\insideskip
\caption{Influence of sample step $t$ and margin $m^0$.}
\label{fig:step_margin}
\outsideskip
\end{figure}

We performed a series of ablative studies to assess the individual contributions of each component in our model and to examine the impact of various hyperparameters. 
The overall results of the ablation study for each component are summarized in \cref{com-ablation}, which indicates that each element contributed to an independent performance improvement.
Additionally, we demonstrate the influence of the sample step in estimating diffusion epistemic uncertainty and the effect of the margin in asymmetric learning, as depicted in \cref{fig:step_margin}.

\textbf{Influence of Diffusion Epistemic Uncertainty} 
As shown in \cref{com-ablation}, integrating diffusion epistemic uncertainty into our model led to the most substantial improvement of 14.2\%/8.0\% ACC/AP compared to the baseline.
The improvement is attributed to the capability of diffusion epistemic uncertainty to capture the general generation properties specific to diffusion models.
However, the detection ACC in the BigGAN subset remained relatively low, indicating the intrinsic differences in generation properties between diffusion-based models and GAN-based models.
% However, the observed improvement in the BigGAN subset is relatively less pronounced, indicating the intrinsic differences in generation properties between diffusion-based models and GAN-based models.

%DEU主要普遍提升diffusion models
%ASL主要提升BigGAN

\textbf{Influence of Asymmetric Learning} 
% The enhancements brought by asymmetric learning mainly occur in subsets produced by the generator that are significantly different from the training set, especially the BigGAN subset.
The improvements brought by asymmetric learning predominantly occurred in subsets generated by models significantly different from those encountered during training. 
Specifically, asymmetric learning led to substantial performance gains on the BigGAN subset, achieving increases of 22.0\% in ACC and 17.6\% in AP over the baseline.
Recent studies~\cite{ojha2023towards} indicate that conventional real-vs-fake classifiers struggle to accurately characterize the real image distribution, causing the real class to act as a ``sink class'' that aggregates features insufficiently similar to the fake features observed during training.
Although diffusion epistemic uncertainty effectively captures intrinsic similarities among diffusion-generated features, it often fails to generalize adequately across diverse generation methods.
In contrast, asymmetric learning facilitates the extraction of generic representations, thereby mitigating the ``sink class'' problem.
By integrating diffusion epistemic uncertainty with asymmetric learning, our proposed approach further improved the detection performance, achieving an overall gain of 15.4\% in ACC and 9.0\% in AP.

\textbf{Influence of Sample Step} As illustrated in \cref{sec:method-deu}, we estimate the diffusion epistemic uncertainty through one-step reconstruction process. 
We set the reconstruction step $t$ to 200 by default following LaRE\textsuperscript{2}.
Moreover, we conducted experiments to verify the impact of $t$ selection on the performance.
The results in \cref{fig:step} demonstrate our method is relatively robust to $t$, and using $t\in[100,400]$ results in comparable performance.

% \begin{figure}[t]
% \centering
% \includegraphics[width=0.7\columnwidth]{pics/step_acc_ap.pdf}
% \caption{Accuracy~(ACC, \%) / average precision~(AP, \%) with reconstruction step $t$ range from 100 to 700. The results indicate that our model is robust to the selection of $t$.}
% \label{fig:step}
% \outsideskip
% \end{figure}

\textbf{Influence of Margin in Asymmetric Learning}
Using a relatively smaller margin for the real class is a crucial aspect of the design of asymmetric contrastive learning. 
To evaluate the impact of different margin values $m^0$ on performance, we performed experiments and present the results in \cref{fig:margin}.
As the value of $m^0$ gradually increased, the average accuracy ACC and AP initially rised, reaching their optimal values at 0.6, and then declined rapidly. 
As discussed earlier, this is because a heavily compressed feature space struggles to capture the wide range of features exhibited by real-class images, while a properly embedded feature space strikes a balance between separability and generalization.

% \begin{figure}[t]
% \centering
% \includegraphics[width=0.7\columnwidth]{pics/margin_acc_ap.pdf}
% \caption{Accuracy~(ACC, \%) / average precision~(AP, \%) with the real-class margin $m^0$ range from 0.0 to 1.0.}
% \label{fig:margin}
% \outsideskip
% \end{figure}

%对于real和fake class使用非对称的margin是非对称学习的关键。
%We evaluated the impact of various m
%结果显示performance先升后降
%这说明将real samples的距离限制在一个相对低的margin

% \textbf{Influence of Loss Weight}
% \subsection{Qualitative Results and Visualizations}
% \section{Discussions} \subsection{Limitations}
\section{Conclusions}
%我们提出使用DEU而非dire来指示图片是否，以克服aleatoric的影响
%我们提出asl进一步改善差异过大的
%对比现有方法，无论是cross generators or cross database，我们的方法都表现出来更优的泛化性
%limitation，我们对于DR variants不能检测
% In this work, we propose a novel framework for detecting diffusion-generated images.
% We find recent works using diffusion reconstruction error as a feature exhibit limited generalizability due to the presence of aleatoric uncertainty inherent in the diffusion process. 
% To address this issue, we propose a novel feature called diffusion epistemic uncertainty. 
% This feature serves as a measure of the deviation of an image from the manifold of diffusion-generated images.
% Additionally, we train a binary classifier by asymmetric learning to further enhance the performance in detecting images from generators that significantly differ from the training set.
% Experiments show that our method yields strong generalizability across various generation methods and databases, outperforming existing methods.

In this work, we propose a novel framework for detecting diffusion-generated images. We observe that recent approaches relying on diffusion reconstruction error as a feature exhibit limited generalizability due to the influence of aleatoric uncertainty inherent in the diffusion process. To address this, we introduce a new feature, diffusion epistemic uncertainty, which quantifies the deviation of an image from the manifold of diffusion-generated images.
% Furthermore, we employ asymmetric learning to train classifiers, enhancing performance in detecting images generated by diffusion models. 
Experimental results demonstrate that our method achieves state-of-the-art generalizability across a variety of generation methods and datasets.

% \raggedbottom

\section*{Acknowledgments}
Bing Bai was supported by the Young Elite Scientists Sponsorship Program by CAST under contract No.~2022QNRC001.

{
    \small
    \bibliographystyle{ieeenat_fullname}
    \bibliography{main}
}

\clearpage
\appendix
% \section{Appendix}
% \subsection{Additional Related Work}

%MLLM for deepfake

%zero-shot for deepfake
% ZeroFake: Zero-Shot Detection of Fake Images Generated and Edited by Text-to-Image Generation Models
% Visual Language Models as Zero-Shot Deepfake Detectors
% Zero-Shot Detection of AI-Generated Images
% Few-Shot Learner Generalizes Across AI-Generated Image Detection

\section{Generalization of DEUA on Diverse Generative Models}
\label{app:gen}
% Although DEUA is primarily designed for diffusion-based detection and is orthogonal to many unified detection methods, it remains competitive with state-of-the-art unified detectors and can be a strong component in broader frameworks.
% To substantiate this, we further evaluated DEUA on the \emph{UniversalFakeDetect} dataset, which covers a wide range of generative models.
% Under the SDv1.4 setting and using publicly available pretrained models or code, DEUA achieved competitive or superior performance when trained on diffusion models (\cref{tab:com-universal}), demonstrating its strong generalizability.
% DEUA also maintains high accuracy on novel diffusion transformers and autoregressive models (\cref{tab:latest}), further confirming its robustness.
Although DEUA is primarily designed for diffusion-based detection and is orthogonal to many unified detection methods, it remains highly competitive with state-of-the-art unified detectors and can serve as a robust component within broader detection frameworks. To comprehensively assess its generalization ability, we further evaluated DEUA on the \emph{UniversalFakeDetect} dataset, which encompasses a diverse set of generative models, including both diffusion and non-diffusion architectures. Under the SDv1.4 setting and leveraging only publicly available pretrained models or code, DEUA achieved competitive or even superior performance compared to existing unified detectors when trained on diffusion models (\cref{tab:com-universal}). These results demonstrate that DEUA not only excels in its targeted domain but also exhibits strong generalizability across a wide spectrum of generative models.

\begin{table*}[!b]
\centering
\resizebox{\linewidth}{!}{%
\begin{tabular}{lcccccccccccccccccccccc}
\toprule
\multirow{3}{*}{Method}
&\multicolumn{6}{c}{GAN}
&\multicolumn{1}{c}{\multirow{3}{*}{\makecell[c]{Deep \\ fakes}}} 
&\multicolumn{2}{c}{Low level}
&\multicolumn{2}{c}{Perceptual loss}
&\multicolumn{1}{c}{\multirow{3}{*}{Guided}} 
&\multicolumn{3}{c}{LDM}
&\multicolumn{3}{c}{Glide}
&\multicolumn{1}{c}{\multirow{3}{*}{Dalle}}
&\multicolumn{1}{c}{\multirow{3}{*}{Avg.}}\\
\cmidrule(lr){2-7} 
\cmidrule(lr){9-10} 
\cmidrule(lr){11-12} 
\cmidrule(lr){14-16} 
\cmidrule(lr){17-19}
&\multicolumn{1}{c}{\makecell[c]{Pro- \\ GAN}}
&\multicolumn{1}{c}{\makecell[c]{Cycle- \\ GAN}}
&\multicolumn{1}{c}{\makecell[c]{Big- \\ GAN}}
&\multicolumn{1}{c}{\makecell[c]{Style- \\ GAN}}
&\multicolumn{1}{c}{\makecell[c]{Gau- \\ GAN}}
&\multicolumn{1}{c}{\makecell[c]{Star- \\ GAN}}
&&\multicolumn{1}{c}{SITD}
&\multicolumn{1}{c}{SAN}
&\multicolumn{1}{c}{CRN}
&\multicolumn{1}{c}{IMLE}
&&\multicolumn{1}{c}{\makecell[c]{200 \\ steps}}
&\multicolumn{1}{c}{\makecell[c]{200 \\ w/cfg}}
&\multicolumn{1}{c}{\makecell[c]{100 \\ steps}}
&\multicolumn{1}{c}{\makecell[c]{100 \\ 27}}
&\multicolumn{1}{c}{\makecell[c]{50 \\ 27}}
&\multicolumn{1}{c}{\makecell[c]{100 \\ 10}}\\
\midrule
NPR (ProGAN)&
99.8& 	95.0 &	87.6 &	96.2 &	86.6 &	99.8 &	76.9 &	66.9 &	98.6 &	50.0 &	50.0 &	84.6 &	97.7 &	98.0 &	98.2 &	96.3 &	97.2 &	97.4 &	87.2 &	87.6  \\
FatFormer (ProGAN)& 
99.9 &	99.3 &	99.5 &	97.2 &	99.4 &	99.8 &	93.2 &	81.1 &	68.0 &	69.5 &	69.5 &	76.0 &	98.6 &	94.9 &	98.7 &	94.4 &	94.7 &	94.2 &	98.8 &	90.9 \\
% C2P-CLIP (ProGAN)& 
% 100.0 &	97.3 &	99.1 &	96.4 &	99.2 &	99.6 &	93.8 &	95.6 &	64.4 &	93.3 &	93.3 &	69.1 &	99.3 &	97.3 &	99.3 &	95.3 &	95.3 &	96.1 &	98.6 &	93.8 \\
NPR (SDv1.4)& 
57.2 &	73.8 &	65.2 &	66.0 &	53.5 &	99.0 &	52.9 &	53.0 &	68.4 &	48.8 &	50.8 &	56.2 &	92.6 &	92.9 &	92.7 &	90.8 &	86.4 &	89.9 &	69.5 &	71.6 \\
DRCT (SDv1.4)& 
99.6 &	93.6 &	87.6 &	99.2 &	90.1 &	99.9 &	72.3 &	67.8 &	60.5 &	68.2 &	59.3 &	92.9 &	99.8 &	99.6 &	99.8 &	99.8 &	99.8 &	99.9 &	91.2 &	88.5 \\
\hline
DEUA (SDv1.4)& 
99.5 &	94.2 &	85.3 &	98.4 &	90.5 &	99.5 &	80.6 &	72.5 &	76.4 &	71.3 &	74.5 &	94.8 &	99.5 &	99.6 &	99.9 &	99.6 &	99.8 &	99.8 &	96.4 &	91.2 \\
\bottomrule
\end{tabular}%
}
\caption{ACC comparisons on the UniversalFakeDetect Dataset.
Results of NPR, FatFormer and C2P-CLIP trained on ProGAN are from paper C2P-CLIP.
Results of NPR and DRCT trained on GenImage SDv1.4 are obtained using their official checkpoints.}
\label{tab:com-universal}
\end{table*}

Furthermore, we investigated the robustness of DEUA on novel generative paradigms, such as diffusion transformers and autoregressive models, which represent the latest advancements in image synthesis. As shown in \cref{tab:latest}, DEUA consistently maintains high detection accuracy on these emerging architectures, further confirming its adaptability and effectiveness. These findings suggest that, despite being tailored for diffusion-based detection, DEUA possesses the flexibility to handle a broad range of generative models, making it a promising candidate for integration into future unified deepfake detection systems.

\begin{table}[ht]
\centering
\resizebox{\linewidth}{!}{%
\begin{tabular}{@{}lccccc@{}}
\toprule
\multirow{3}{*}{Method}
&\multicolumn{1}{c}{Unet}
&\multicolumn{2}{c}{Transformer}
&\multicolumn{1}{c}{Autoregressive}
&\multicolumn{1}{c}{\multirow{3}{*}{Avg.}}\\
\cmidrule(lr){2-2} 
\cmidrule(lr){3-4}
\cmidrule(lr){5-5}
&SDv1.4&SDv3&SDv3.5&JanusPRO\\
\midrule
NPR~(ProGAN)& 76.6&	76.2&	77.8&	76.3&	76.7 \\
FatFormer~(ProGAN)& 83.2 & 70.1& 65.4& 82.6& 75.3\\
NPR~(SDv1.4)& 
98.2&	80.1&	83.6&	86.5&	87.1
\\
% LaRE &  
% 100.0&	80.8&	78.4&	86.9&	86.5 \\
DRCT~(SDv1.4) &  
95.1&	91.2&	90.4&	93.9&	92.7 \\
\hline
DEUA~(SDv1.4) & 
99.2&	97.3&	96.1&	98.1&	97.7 \\
\bottomrule
\end{tabular}
}
\caption{ACC comparison on new generators. SDv3, sdv3.5 and JanusPRO are collected following GenImage.
Results of NPR, FatFormer and DRCT are obtained using their official checkpoints.}
\label{tab:latest}

\end{table}

\end{document}